\documentclass[11pt]{article}

\usepackage[final]{acl}

\usepackage{times}
\usepackage{latexsym}
\usepackage{tcolorbox}
\usepackage[T1]{fontenc}

\usepackage[utf8]{inputenc}

\usepackage{microtype}

\usepackage{inconsolata}

\usepackage{amsmath}
\usepackage{cleveref}

\usepackage{graphicx}
\usepackage{multirow}
\usepackage{booktabs}
\usepackage{amsfonts}
\usepackage{caption,subcaption}
\usepackage{soul}
\usepackage{booktabs}
\usepackage{tabularx}
\usepackage{algorithm}
\usepackage{algpseudocode}
\usepackage{enumitem}
\newcommand{\ensuretext}[1]{#1}
\newcommand{\nertcomment}[4]{\ensuretext{\textcolor{#3}{[\ensuretext{\textcolor{#3}{\ensuremath{^{\textsc{#1}}_{\textsc{#2}}}}} #4]}}}
\newcommand{\sw}[1]{\nertcomment{S}{W}{orange}{#1}}
\newcommand{\ab}[1]{\nertcomment{A}{B}{blue}{#1}}

\title{For What Reason?\\ Interpreting Models' Encoding of Causation and Antithesis}

\author{Abhidip Bhattacharyya \\
  University of Massachusetts Amherst \\
  \texttt{abhidipbhatt@umass.edu} \\\And
  Shira Wein \\
  University of South Florida \\
  \texttt{shirawein@usf.edu} \\}

\begin{document}
\maketitle
\begin{abstract}
    Discourse relations provide document structure, critical to language understanding and enabling language model performance and ethicality.
In this work, we investigate how instruction-tuned Transformer models (LLaMA and Mistral) encode discourse relations in English, with a particular focus on the contrasting relations of causation and antithesis. Framing the task as a next-token prediction task and applying a suite of interpretability techniques to test model internals, our findings show that certain early layers make predictive decisions at mid-sequence tokens, while some mid-level layers finalize their decisions closer to the last token. Most of the remaining layers primarily propagate earlier decisions rather than actively influencing them. Additionally, we observe that some layers exhibit a preference for one answer over alternatives, suggesting asymmetric representation of discourse-based reasoning.\footnote{Our code is available at \url{https://github.com/abhidipbhattacharyya/causation_vs_antithesis}}
\end{abstract}

\section{Introduction}
    \label{sec:intro}

Discourse relations are essential for natural language understanding, characterizing the structure of sentences within a document. Additionally, discourse structure is a critical component of enabling language model alignment with human preferences and promoting ethical behavior \citep{harrison-etal-2019-maximizing,husunbeyi-etal-2022-identifying,adewoyin-etal-2022-rstgen,kim2025align,nakshatri-etal-2025-talking}.
However, limited prior work has investigated discourse encoding within Transformers, with only one recent work has investigated discourse encoding in Transformers using Transformer circuit discovery to identify ``discursive circuits'' \citep{miao-kan-2025-discursive}.



\begin{figure}[t]
\begin{subfigure}[t]{0.45\textwidth}
\centering
\begin{tcolorbox}[
  colback=blue!5!white,      
  colframe=blue!75!black,    
  coltitle=white,            
  boxrule=0.5pt,
]
    \textbf{Causation}: John studied hard, so he passed the exam.\\
    \textbf{Antithesis}: John studied hard, yet he failed the exam.
\end{tcolorbox}
\vspace{-0.3cm}
\caption{Examples illustrating causal and antithetical relationships. Causation links an action to its expected outcome, whereas antithesis presents a contrasting outcome.}
\label{subfig:antithesis}
\end{subfigure}
\begin{subfigure}[t]{0.45\textwidth}
\vspace{0.4cm}
\centering
 \begin{tcolorbox}[
        colback=gray!10,      
        colframe=black!50,    
        coltitle=black,            
        boxrule=0.5pt,
        ]
    \textbf{Causation}: John studied hard, \textit{so} he was \textit{able} to pass the exam.\\
    \textbf{Antithesis}: John studied hard, \textit{yet} he was \textit{unable} to pass the exam.
    \end{tcolorbox}
    \vspace{-0.3cm}
    \caption{The same examples rewritten using the lexical contrast between \textit{able} and \textit{unable}. The conjunction ``so'' signals causation, while ``yet'' marks an antithetical relationship. Apart from the four highlighted tokens, the two sentences are identical.}
\label{subfig:afterwards}
\end{subfigure}
\end{figure}

In this work, we investigate Transformer encoding of discourse-level phenomena in English with a focus on two frequent, and oppositional relations: causation and antithesis. In Rhetorical Structure Theory \citep[RST;][]{rst1, rst2}, causation represents the 
relationship between an action and its consequence. In contrast, the rhetorical figure of antithesis employs parallel 
grammatical structures to juxtapose contrasting or opposing ideas. An illustrative example of these discourse relations is shown 
in \Cref{subfig:antithesis}. Within our  investigation of discourse encoding within Transformer models, our research questions include:

\begin{itemize}
  \item[RQ1.] Which token-level features drive the model's decisions between antithesis and causation?
  \item[RQ2.] Which layers play a pivotal role in identifying antithesis and causation?
  \item[RQ3.] What key concepts help the model distinguish between antithesis and causation?
  \item[RQ4.] How does discursive information propagate across layers within the model?
\end{itemize}

To address these research questions, like \citet{miao-kan-2025-discursive}, we formulate the task as a next token prediction task, where the token indicates either a causal or antithetical relationship identified within the sentence. 
We perform a range of causal interventions and apply techniques from automated circuit discovery. Our results show that early and mid layers encode local relational meanings signaled by discourse markers such as ``so''\slash``yet,'' while higher layers primarily propagate or refine these signals. 
In contrast, interventions at the last token indicate that lower layers integrate sentence meaning, whereas higher layers make and maintain the final decision. We also observe that certain layers exhibit an inherent bias toward one option over the other. 
At the circuit level, we find a core set of stable connections that consistently drive decision making and information flow, although specific edges involved may vary depending on verb type and the number of demonstrations.
Our findings provide insight into how critical discourse relations are encoded in Transformer models, enabling heightened model performance and alignment to human preferences. 

\section{Methods \& Experiments}
    \label{sec:methods}
        
In this section, we detail the task formalism (\Cref{ssec:task}), data preparation procedures (\Cref{ssec:data}), model configurations (\Cref{ssec:models}), and interpretability techniques (\Cref{ssec:methods}).

\subsection{Task}
\label{ssec:task}

We prompt models to complete the answer in a way that indicates a causation or antithesis within the sentence. To restrict the model prediction space we design
the answer to be either ``able'' or ``unable.''
We focus on decoder-only models, namely LLaMA 3 \citep{dubey2024llama} and Mistral \citep{jiang2023mistral7b}, using their 
instruction-tuned variants with 7 billion parameters. The task is formulated as a next-token prediction problem, allowing us to probe how these models internally represent and differentiate between distinct discourse relations.

We formulate the task of differentiating between causation and antithesis as a next-token prediction problem. In our setup, each sentence is constructed to include either ``able'' or ``unable.'' For example, the sentences shown in \autoref{subfig:antithesis} are reformulated as illustrated in \autoref{subfig:afterwards}.
      
To prompt the LLM for this task, we replace `able'' or unable'' with a blank and ask the model to predict the missing word, instructing it to choose only between able'' and unable.'' The prompt includes few-shot examples (two per class) to clarify the task.
We initially find with 1–6 shot prompting and found that 4-shot prompting produced more consistent results. Therefore, experiments in this paper follows 4 shots ((two per class).
However, subsequent experiments show that 0-shot prompting achieved the best overall performance. \autoref{fig:llama_shots} in \autoref{appendix_a} presents prediction accuracy across different numbers of demonstrations, and an example prompt is shown in \autoref{fig:prompt}. 
Additional details on prompt design are provided in \autoref{appendix_a}.
    \begin{figure}[ht]
        \centering
        \begin{tcolorbox}[
            colback=gray!10,      
            colframe=black!50,    
            coltitle=black,            
            boxrule=0.5pt,
        ]
        \textbf{User Prompt: } 
        1. Sentence: John got the train this morning, so he was \_\_\_ to reach on time.\\
        Answer: able\\
        2. Sentence: The team practiced all week, so they were \_\_\_ to perform well.\\
        Answer: able\\
        3. Sentence: The team practiced all week, yet they were \_\_\_ to perform well.\\
        Answer: unable\\
        4. Sentence: Lena studied hard, yet she was \_\_\_ to pass the exam.\\
        Answer: unable\\

        Now complete the following:\\
        5. Sentence: He slept well last night, yet he was \_\_\_ to run this morning.\\
        Answer:
        \end{tcolorbox}
        \vspace{-0.3cm}
        \caption{Example of our few-shot prompt. Our system prompt can be found in \Cref{fig:sys_prompt} in \autoref{appendix_a}.}
        \label{fig:prompt}
    \end{figure}
\subsection{Data}
    \label{ssec:data}

To probe the models' representation of causation and antithesis, we build a small dataset of sentences containing antithesis and their corresponding causation relations. 
We impose a constraint that both sentences in each pair must have the same word count. This constraint is motivated by our planned causal intervention experiments (\Cref{ssec:methods}).

To generate the dataset, we prompt ChatGPT 4o\footnote{Accessed May 2025.}, by first manually creating a few example pairs that illustrate the contrast between causation and antithesis. These examples are then used to prompt ChatGPT to generate additional sentence pairs following the same pattern. We generate 130 such samples to be used to train the probes. Additionally we create 30 samples as our test set, which are manually checked by the authors to ensure that they follow the desired format.

In the initial setup, ``so'' cues ``able'' indicating causation, while ``yet'' cues ``unable'' indicating antithesis. To make the task more challenging, we introduce negative verbs in the satellite clause, which reverses the association between ``able'' and ``unable'' (\Cref{fig:negverb}). 
For such verbs, ``unable'' typically fits causal contexts, whereas ``able'' is more appropriate for antithetical ones. 
This prevents the model from solely relying on cue words ``so'' or ``yet'' to distinguish between causation and antithesis. We suspect that the sentiment of the verb has an impact on the ability of the model to predict either ``able'' or ``unable.'' For example, in \Cref{subfig:afterwards}, the verb sense is positive, whereas in \Cref{fig:negverb}, the verb sense is negative.

    \begin{figure}[htb]
        \centering
            \begin{tcolorbox}[
                colback=orange!15!white,      
                colframe=orange!75!black,    
                coltitle=white,            
                boxrule=0.5pt,
            ]
                \textbf{Causation}: She breaks rules, so she was \textit{unable} to join the team.\\
                \textbf{Antithesis}: She breaks rules, yet she was \textit{able} to join the team.
            \end{tcolorbox}
            \vspace{-0.3cm}
        \caption{An example illustrating how ``unable'' and ``able'' are used to differentiate causation and antithesis. Note that in \autoref{subfig:afterwards}, ``able'' appears in the causation example, whereas ``unable'' is used in the antithesis. This example demonstrates the opposite usage, highlighting how verb semantics can affect interpretation.}

        \label{fig:negverb}
    \end{figure}

Moreover, verb semantics can shift under explicit negation. Therefore, for each case, we also generate examples with explicit negation. Introducing negation often reverses the use of `able' and `unable' in both causation and antithesis cases. For instance, the examples in \Cref{fig:negverb} are flipped in \Cref{fig:notflip} when `not' is introduced in the satellite clause. We then examine model performance with these negated examples. 


\begin{figure}[htb]
        \centering
            \begin{tcolorbox}[
                colback=green!10!white,      
                colframe=green!50!black,     
                coltitle=black,              
                boxrule=0.5pt,
            ]
                \textbf{Causation}: She did not break rules, so she was \textit{able} to join the team.\\
                \textbf{Antithesis}: She did not break rules, yet she was \textit{unable} to join the team.
            \end{tcolorbox}
            \vspace{-0.3cm}
        \caption{An example showing how ``able'' and ``unable'' switch roles when the satellite verb is negated. Examples correspond to those in \autoref{fig:negverb} and their negated forms.}

        \label{fig:notflip}
    \end{figure}

\subsection{Models}
    \label{ssec:models}
We prompt two LLMs of comparable size and from the same architectural family in order to observe common patterns.
Our primary model for investigation is \textbf{LLaMA} \citep{dubey2024llama}, specifically \texttt{LLaMA-3-8B-Instruct}, an instruction-tuned version of LLaMA with 8 billion parameters, with 32 layers and 32 attention heads \citep{llama3modelcard}. 
LlaMa uses grouped query attention with 8 key-value groups.
We use \textbf{Mistral} as our second model~\citep{jiang2023mistral7b}, specifically \texttt{Mistral-7B-Instruct-v0.2} \citep{MistralModelcard}, an instruction-tuned version of Mistral. Similar to LLaMA, Mistral has 32 layers and 32 attention heads. 
However, Mistral employs sliding window attention \citep{longformer,bigbird}.

We opt for instruction-tuned models given our use of modeling prompting in the experimental pipeline. Prior work has shown that instruction tuning does change LLM behavior, such as by altering the self-attention heads to attend more to verbs that appear commonly in prompts versus common verbs \citep{wu-etal-2024-language}, making it unclear how our findings generalize to other foundation models.
\begin{figure*}[htb]
    \centering 
\begin{subfigure}{0.45\textwidth}
  \includegraphics[height=\linewidth]{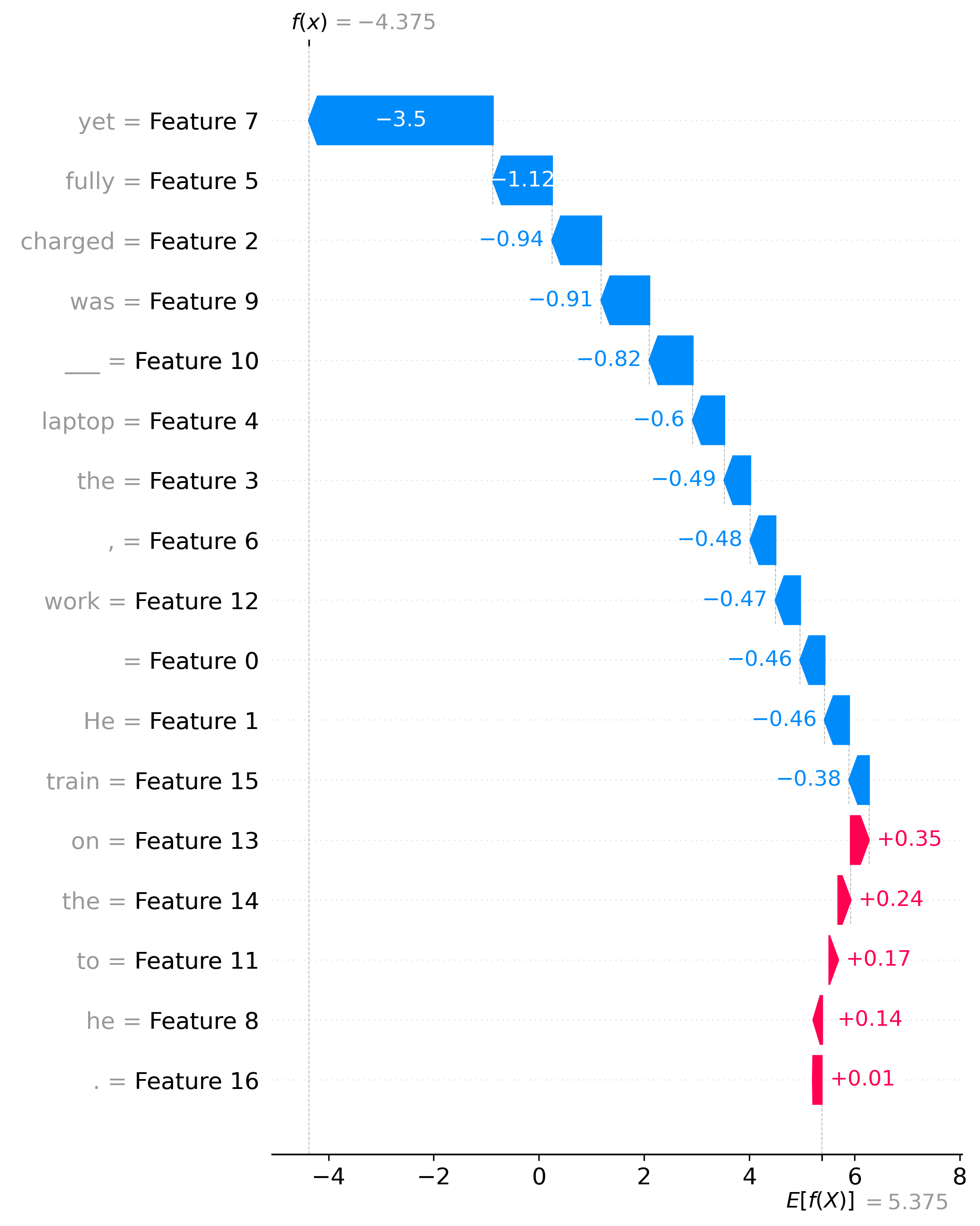}
 \caption{}
  \label{subfig:shapex1}
\end{subfigure} 
\hfill
\begin{subfigure}{0.45\textwidth}
  \includegraphics[height=\linewidth]{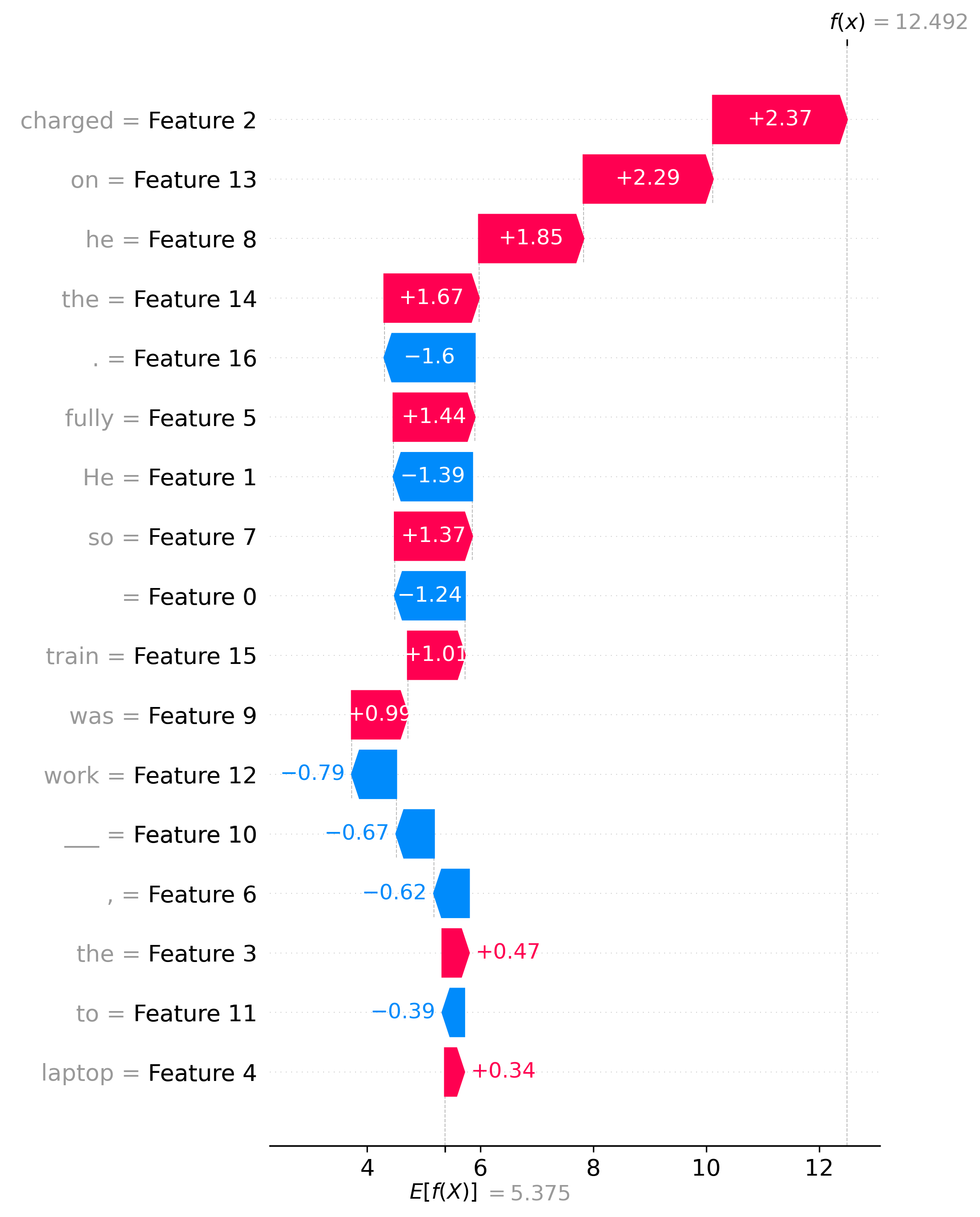}
  \caption{}
  \label{subfig:shapex2}
\end{subfigure} 

\caption{Normalized SHAP contributions for tokens across examples in LLaMA. 
  Subfigures (a) and (b) illustrate example-level SHAP values for individual sentences. }
\label{fig:shap2}
\end{figure*}
\begin{figure*}[htb]
    \centering 
\begin{subfigure}{0.40\textwidth}
  \includegraphics[width=\linewidth]{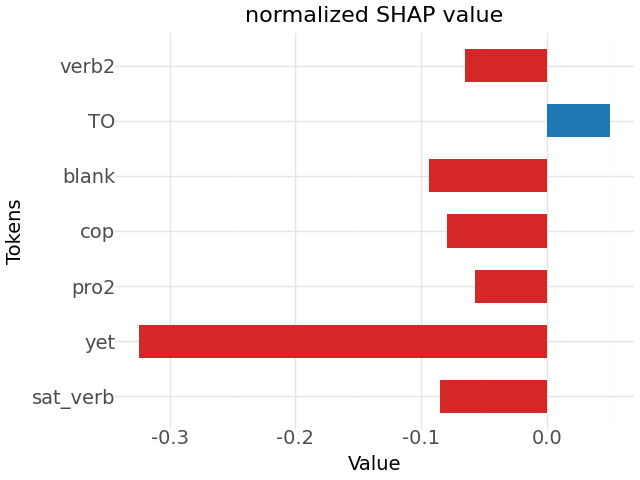}
  \caption{}
  \label{subfig:shap_yet_nor}
\end{subfigure}\hspace{0.15cm} 
\begin{subfigure}{0.40\textwidth}
  \includegraphics[width=\linewidth]{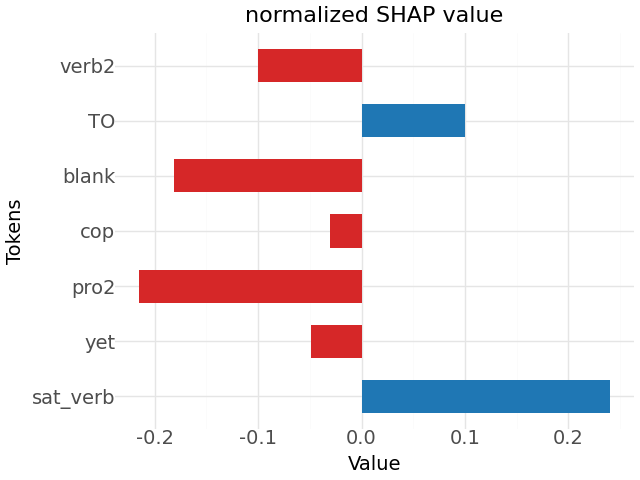}
  \caption{}
  \label{subfig:shap_yet_neg}
\end{subfigure}\hspace{0.15cm} 

\medskip

\begin{subfigure}{0.40\textwidth}
  \includegraphics[width=\linewidth]{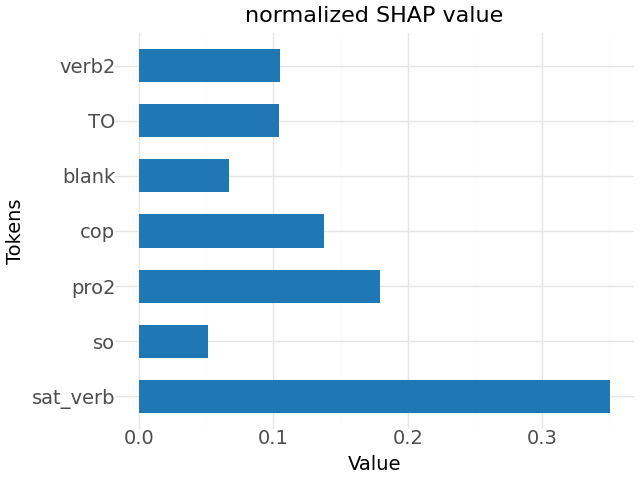}
  \caption{}
  \label{subfig:shap_so_nor}
\end{subfigure}
\begin{subfigure}{0.40\textwidth}
  \includegraphics[width=\linewidth]{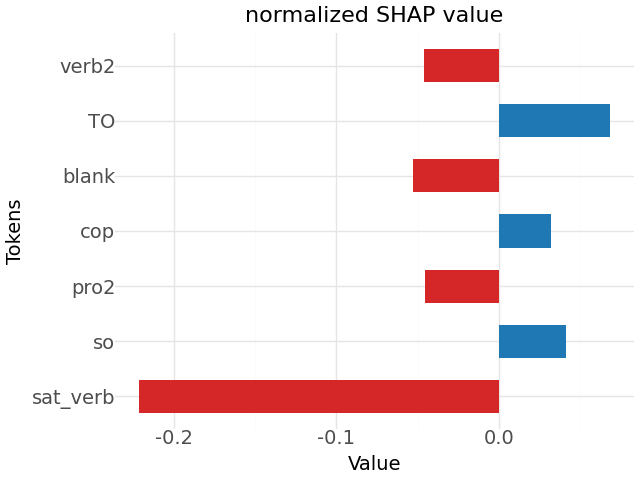}
  \caption{}
  \label{subfig:shap_so_neg}
\end{subfigure}
\caption{Normalized SHAP contributions for tokens across examples in LLaMA. The top row shows results for sentences containing the discourse marker ``yet,'' and the bottom row shows results for sentences containing ``so.'' 
 Subfigures (a) and (c) present the average normalized SHAP values for positive verbs, while 
 Subfigures (b) and (d) present the average normalized SHAP values for negative verbs.}
\label{fig:shap1}
\end{figure*}
\subsection{Interpretability Methods}
\label{ssec:methods}

To address our research questions, we construct several techniques based in mechanistic interpretability. 
The first interpretability technique we operationalize is \textbf{Activation patching} \citep{causal_Geiger,rome_meng,zhang2024towards}, which is a technique used to understand the causal effect of a specific internal representation on a model's predictions. In this approach, a portion of the model's internal representation is replaced with the representation generated from the model's processing of an alternative input. The causal role of the targeted unit is then assessed by analyzing the resulting changes in the output logits (i.e., the model's unnormalized scores for each possible output token). If substituting the representation shifts the logits or probits toward a different output, it indicates that the inspected unit plays a causal role in the model's prediction.

In our activation patching approach, we first run the model on a data instance $x_{base}$ (either antithesis or causation) and measure the logit difference between the two answer choices, ``able'' vs ``unable'', which we denote as $LD(x_{base})$. Second, we run the model on the corresponding counterexample $x_{source}$. The counterexample is created either by switching between ``so''\slash``yet'', or by negating the verb in the satellite. Therefore, the counterexample for antithesis will be causation, and vice versa. We then measure the logit difference between the two answer choices, denoted as $LD(x_{source})$. Next, we run the model on $x_{base}$, but for a target layer and token position, we replace the activation($h_{l,t}$) with the corresponding activation from the $x_{source}$ run. We measure the logit difference as before, denoted by $LD(x_{base}; h_{l,t}^{base} \leftarrow h_{l,t}^{source})$.
Following previous work \citep{ioi,li2025how}, we measure the \textit{normalized logit difference} (NLD):
    \begin{equation}
        \centering  
            NLD = \frac{LD(x_{base}; h_{l,t}^{base})-LD(x_{base})}{LD(x_{source}) -LD(x_{base})}
    \end{equation}

Second, we use \textbf{Logits Attribution} to assess the influence of internal layers in decision making. We measure the alignment of the layer representation with the \emph{logit difference direction} \citep{ioi,logitatt}. 
Let the unembedding matrix be denoted as 
$W_{\text{unemb}} \in \mathbb{R}^{d\times |V|},$
where $|V|$ is the vocabulary size and $d$ is the hidden dimension. For a hidden state representation $h_{L,t} \in \mathbb{R}^d$ at the $L^{th}$ layer and time step \(t\), the logits are given by
$\ell = W_{\text{unemb}}^\top h_{L,t} ,$ which is a vector in \(\mathbb{R}^{|V|}\).
However, we are specifically interested in the logit difference between the correct answer (\texttt{ca}) and an incorrect answer (\texttt{ia}). This can be expressed as
\begin{equation}
\ell_{\text{ca}} - \ell_{\text{ia}} 
= \big(W_{\text{unemb}}[\texttt{ca}] - W_{\text{unemb}}[\texttt{ia}]\big)^\top h_{L,t}.
\end{equation}
 $W_{\text{unemb}}[\texttt{ca}]$ and $W_{\text{unemb}}[\texttt{ia}]$ denote the column vectors of the unembedding matrix for the correct and incorrect tokens. Thus, the logit difference reduces to the dot product between the hidden state representation $h_{L,t}$ and the \emph{logit difference direction}. A higher dot product indicates a greater affinity of the layer toward the correct answer.

We also conduct three \textbf{probing} experiments \citep{probing2,probing1} on the internal representations of a frozen language model, where hidden states are extracted layer-wise, and a logistic regression classifier is trained as a probe:

\begin{itemize}[noitemsep]
    \item \textbf{``Able'' vs. ``unable''}: Layer-wise token representations are used to train a binary classifier that predicts whether the model's output corresponds to the answer \emph{able} or \emph{unable}.
    
    \item \textbf{Verb sentiment in the satellite} (sat-verb-sentiment): The probe is trained to classify the sentiment polarity (positive vs. negative) of the verb in the satellite clause (e.g., the verb in \autoref{subfig:afterwards} is positive, while that in \autoref{fig:negverb} is negative).
    
    \item \textbf{Overall sentiment of the satellite} (sat-clause-sentiment): The probe predicts the overall sentiment of the satellite clause, accounting for the effect of negation. For instance, the satellite in \autoref{subfig:afterwards} is positive, but negation reverses its polarity; conversely, in \autoref{fig:negverb} the clause is negative, but negation yields a positive sentiment.
\end{itemize}

Finally, to analyze information flow and quantify edge contributions, we employ \textbf{attribution patching} \citep{eap2,eap1,eap3}. While \emph{activation patching} provides a more precise intervention \citep{causal_Geiger,rome_meng,zhang2024towards}, it is computationally expensive when identifying relevant subgraphs within the computation graph~\citep{acdc,eap1}.
Attribution patching offers a more efficient alternative by approximating the effect of activation patching using a first-order Taylor expansion of the evaluation function around a base run, enabling scalable estimation of edge importance \citep{eap1,eap3}.
 \begin{multline}
 L(x_{\text{base}} \mid \text{do}(E = e_{\text{source}})) 
 \approx L(x_{\text{base}}) \\
 + \underbrace{(e_{\text{source}} - e_{\text{base}})^{\top} 
 \frac{\partial}{\partial e_{\text{base}}} 
 L(x_{\text{base}} \mid \text{do}(E = e_{\text{base}}))}_{\Delta_e L \; \text{(attribution score)}}
 \tag{2}
     \label{eq:attri_eq}
 \end{multline}
Following \cite{eap1}, we compute absolute attribution scores and use them as measures of edge importance. We then select the \emph{top-$k$} edges based on these scores.

\subsection{Computational Resources}
We run all experiments on a single NVIDIA Titan RTX GPU (24 GB VRAM). 
Edge attribution patching experiments are conducted on a single NVIDIA A100 GPU (80 GB VRAM) due to their higher memory requirements.
\section{Results and Observations}
    The models perform well on our ``able'' versus ``unable'' task, with LLaMA achieving a zero-shot F1 score of 0.96, and Mistral achieving a zero-shot F1 score of 0.80.
This high-performance enables effective interpretation of the models' encoding of discourse relations.
We now present our experimental results addressing our research questions.

\subsection{RQ1: Which token-level features influence the model's decisions?}
To investigate which tokens have the greatest influence on the model's antithesis versus causation prediction, we first analyze the log-odds of the model's predictions using SHAP \citep{shap}. \Cref{subfig:shapex1} and \Cref{subfig:shapex2} 
present example-level SHAP values for individual sentences. 
\Cref{subfig:shap_yet_nor} and \ref{subfig:shap_so_nor} show the average normalized average SHAP values for sentences containing ``yet'' and ``so'' paired with positive verbs, while \Cref{subfig:shap_yet_neg} and \ref{subfig:shap_so_neg} show the corresponding values for sentences with negative verbs.

Notably, as illustrated in \Cref{subfig:shap_so_nor,subfig:shap_yet_neg}, the verb contributions are aligned: for positive verbs, ``so'' predominantly drives the decision towards ``able,'' whereas for negative verbs, ``yet'' has a similar effect. 
Conversely, in Figure~\ref{subfig:shap_yet_nor} and \ref{subfig:shap_so_neg}, ``yet'' with positive verbs and ``so'' with negative verbs tend to lead to ``unable'' as the expected response. This supports our intuition regarding the influence of verb type at higher-level causal abstraction, as described in \Cref{ssec:data}.

To quantify these differences, we perform a two-sample t-test across combinations of verb types and discourse markers—for example, comparing positive sentiment verbs with ``so'' and negative sentiment verbs with ``yet'' (\autoref{fig:shap1}e and \autoref{fig:shap1}c), as well as positive verbs with ``yet'' and negative sentiment verbs with ``so'' (\autoref{fig:shap1}b and \autoref{fig:shap1}f). The results indicate that the group means are significantly different, demonstrating that although the direction of the average contribution of ``so'' and ``yet'' aligns across verb types, their impact on model decisions is not identical. Depending on the verb type, specific tokens have a stronger influence in determining the output.\footnote{A detailed summary of t-test results for the verb position across all configurations is provided in \autoref{tab:ttestVerb}.}

Guided by these findings, we select our intervention locations at the ``so''\slash ``yet'' token for the remaining research questions.
Additionally, we choose the last token position, as higher-level semantic information is likely encoded there \citep{ioi, WiegreffeTBHS25}.

\begin{figure*}[htb]
    \centering 
\begin{subfigure}{0.36\textwidth}
  \includegraphics[width=\linewidth]{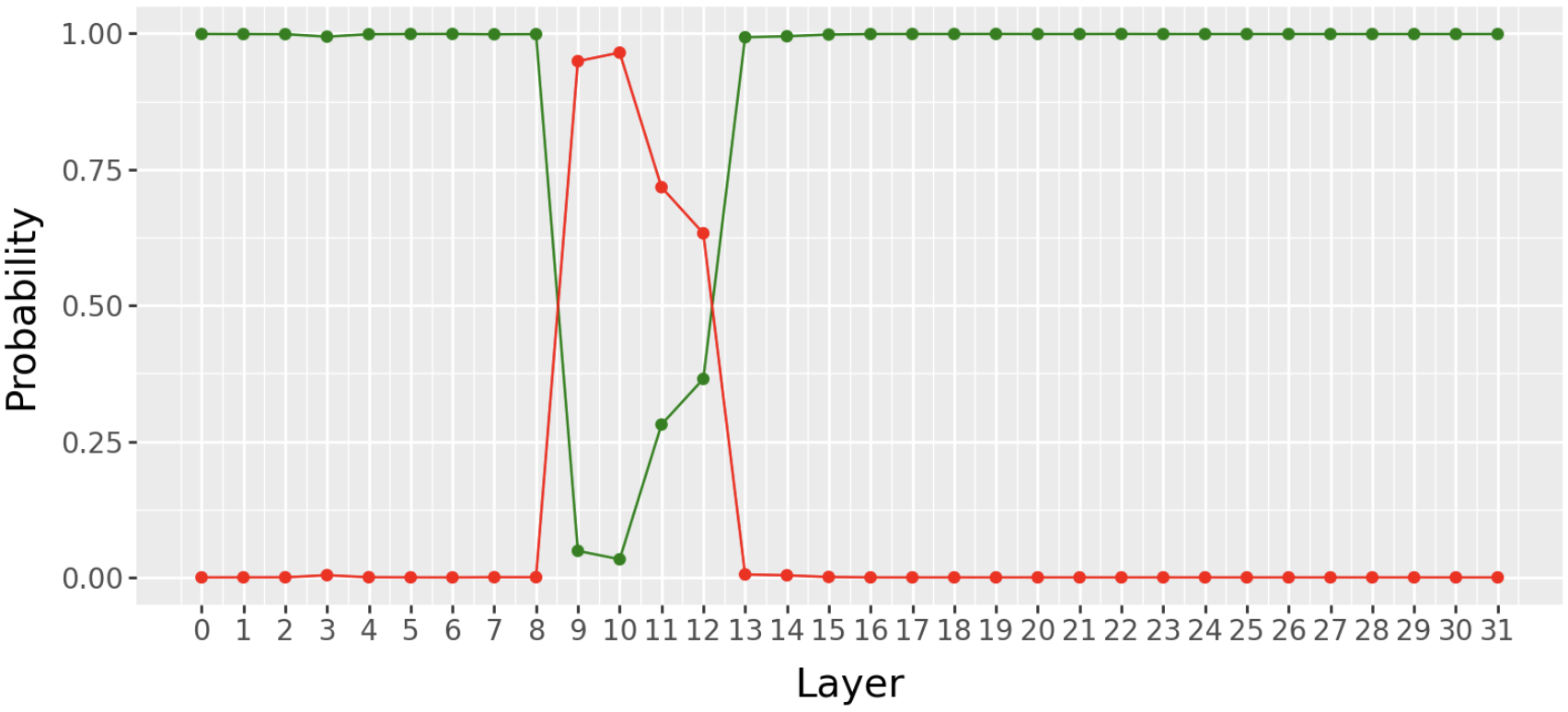}
  \caption{}
  \label{subfig:actpatching12}
\end{subfigure}\hspace{0.15cm} 
\begin{subfigure}{0.46\textwidth}
  \includegraphics[width=\linewidth]{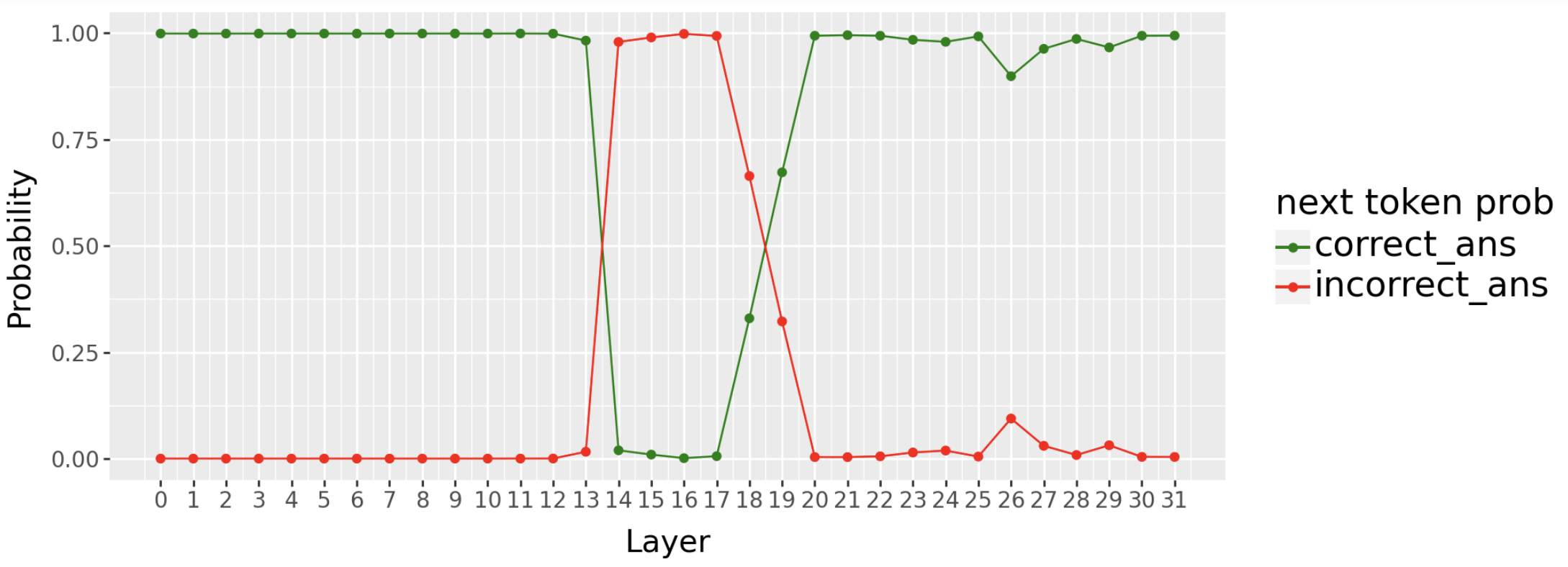}
  \caption{}
  \label{subfig:actpatching15}
\end{subfigure} 

\medskip

\begin{subfigure}{0.36\textwidth}
  \includegraphics[width=\linewidth]{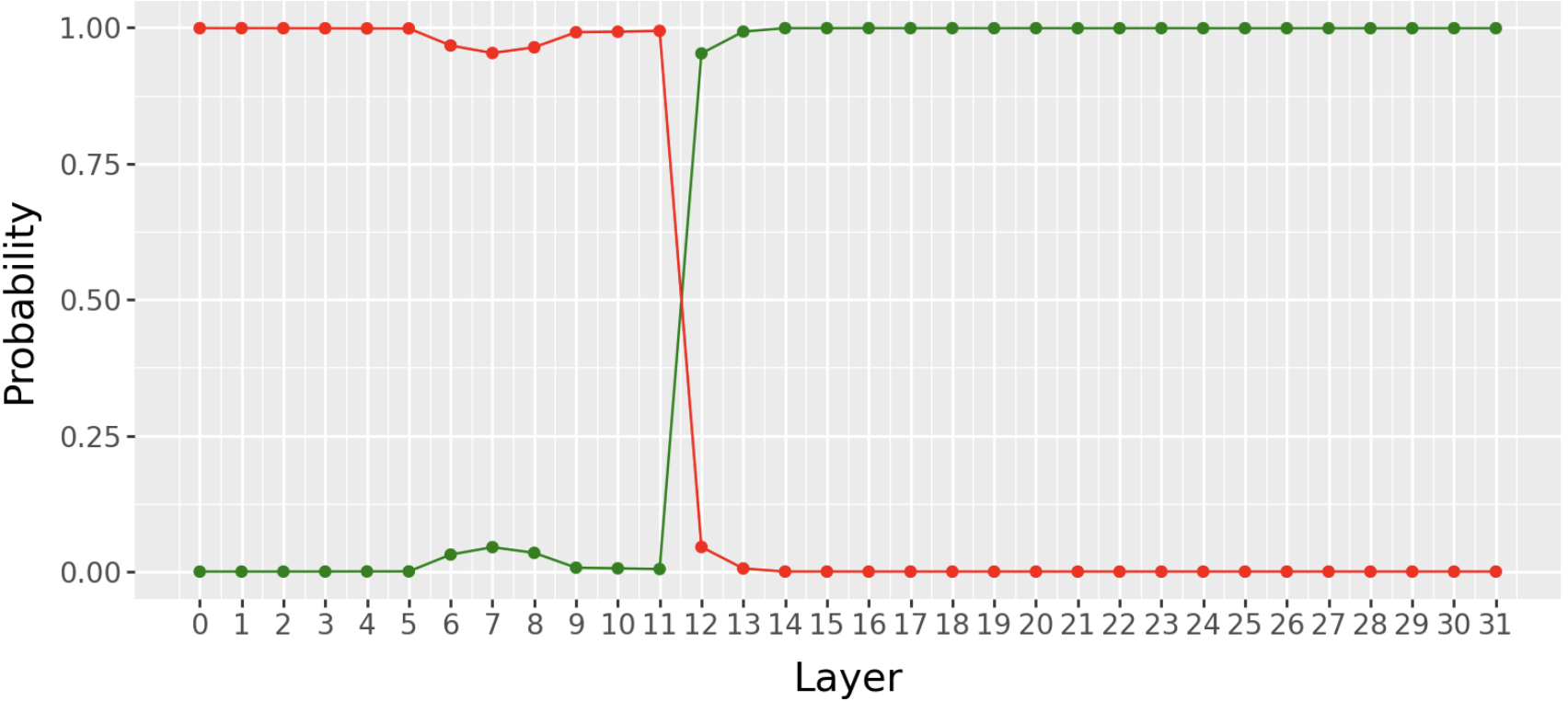}
  \caption{}
  \label{subfig:actpatching13}
\end{subfigure}
\begin{subfigure}{0.46\textwidth}
  \includegraphics[width=\linewidth]{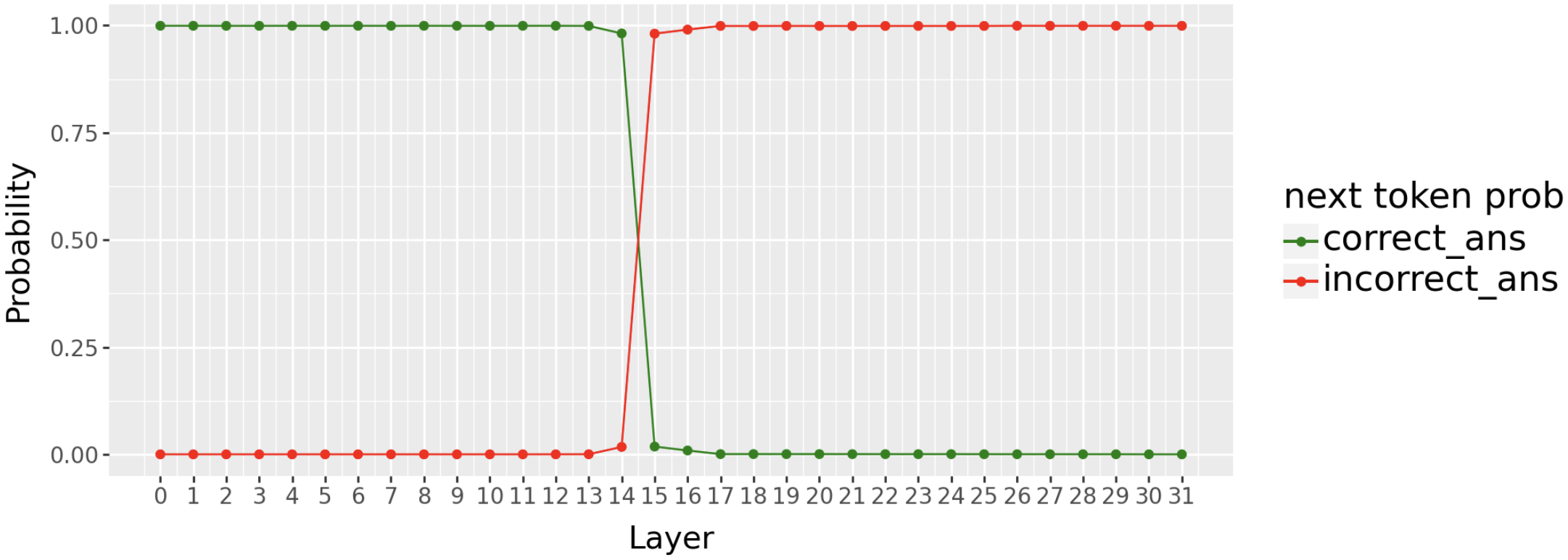}
  \caption{}
  \label{subfig:actpatching16}
\end{subfigure}
\caption{Change in probability from the correct answer (``able'' or ``unable'') to the wrong answer under activation patching between causation and antithesis with a positive verb.
The left column shows the change in probability when patched at the token location for ``so'' and ``yet,'' while the right column shows the same change at the last token location. \ref{subfig:actpatching12} and \ref{subfig:actpatching15} show the effect of patching at the attention output, and \ref{subfig:actpatching13} and \ref{subfig:actpatching16} depict the effect of patching at the block output.}
\label{fig:actpatching1}
\end{figure*}

\subsection{RQ2: Which layers play a pivotal role in decision making?}
To understand which layers play a pivotal role in distinguishing antithesis and causation, we patch activations at the token 
position corresponding to ``so''\slash``yet'' and at the final token position.

\Cref{fig:actpatching1} illustrates the results of activation patching in the LLaMA model. \Cref{fig:mistral_actpatching} in \autoref{appendix_a} depicts the same for the Mistral model.
We observe that MLP layer activations are largely insensitive to the patching intervention(\Cref{fig:mlp_patching}). In contrast, 
attention and block outputs display meaningful patterns. Specifically, patching at the ``so''\slash``yet'' token position in 
early-to-mid attention layers transiently alters the output probits, whereas patching at the final token position 
affects probits in later mid layers. Notably, in both cases, interventions at higher layers fail to sustain this 
flipping effect. Similar trends are observed in the block outputs (\autoref{subfig:actpatching13}).

This flipping of model output at early-mid layers for ``so''\slash``yet'' tokens suggests these layers encode 
local relational meanings (e.g., causal vs. contrastive) signaled by discourse markers. 
In contrast, higher layers appear to primarily propagate or refine the decision rather than reevaluate it, hence perturbations at this stage don't flip the output. 
Conversely, the sensitivity to patching at the last token position in the mid layers indicates that these layers integrate sentence-level meaning before producing logits. 
Once this integration is complete, later layers merely transmit the decision, rendering further interventions ineffective.

Interestingly, when patching is applied at the block output of the final token, the lower layers remain relatively 
stable, but the model's decision flips around layer 15. This aligns with our observations from RQ1: once mid-layer activations 
consolidate the model's decision, higher layers play a passive role, and swapping their activations with those from corrupted inputs cannot reverse the output.
Similar observations are made when negating the satellite verb; see \Cref{fig:patching_with_NOT_Ans_unable,fig:patching_with_NOT_Ans_able} in \autoref{appendix_a}.

\begin{figure*}[t]
    \centering 
\begin{subfigure}{0.45\textwidth}
  \includegraphics[width=\linewidth]{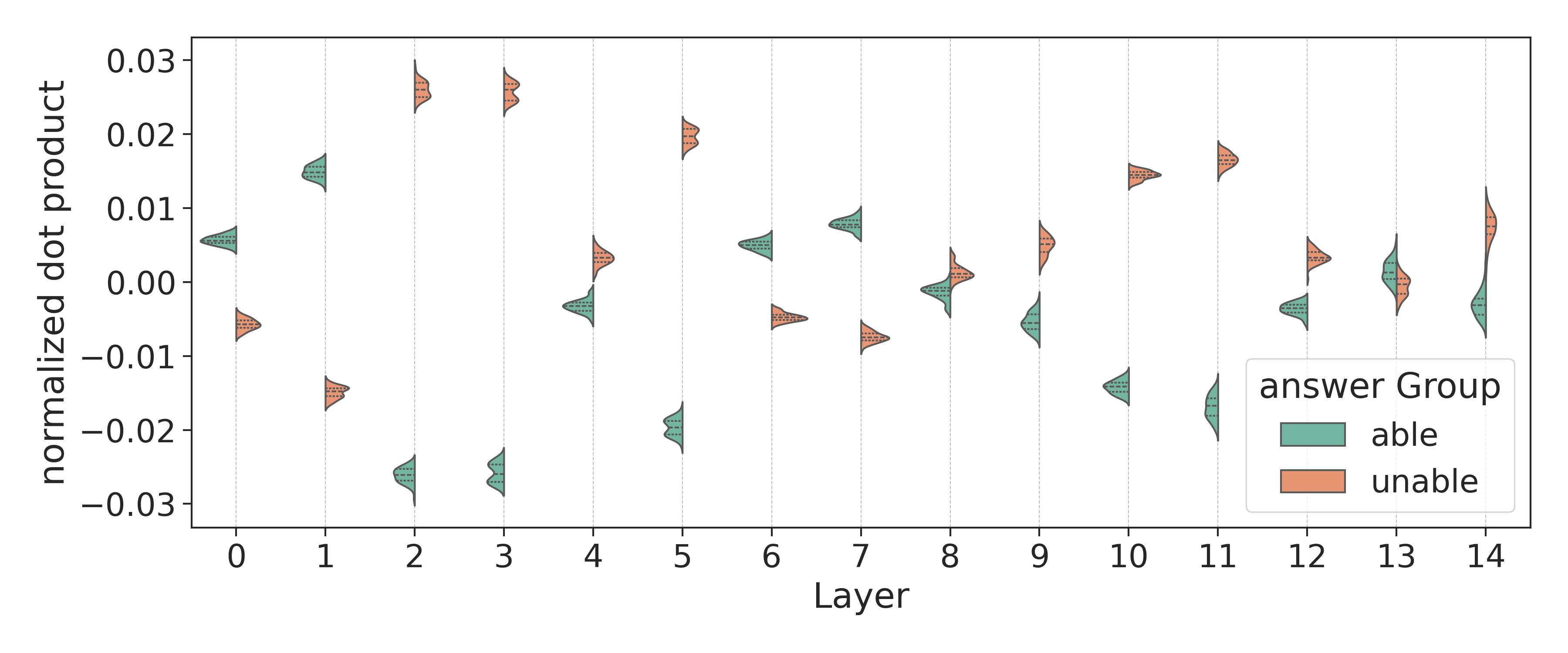}
  \caption{}
  \label{subfig:logatt1}
\end{subfigure}\hspace{0.15cm} 
\begin{subfigure}{0.45\textwidth}
  \includegraphics[width=\linewidth]{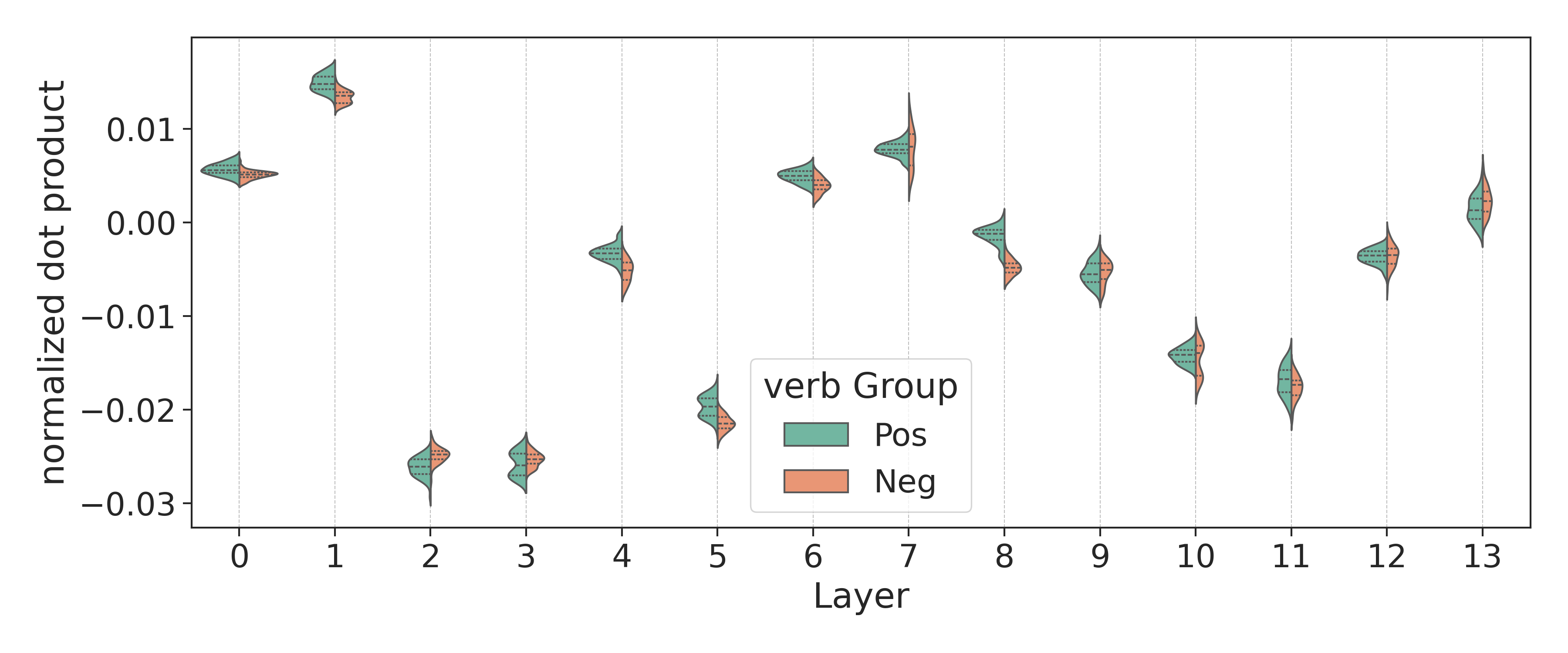}
  \caption{}
  \label{subfig:logatt2}
\end{subfigure}\hspace{0.15cm} 
\medskip
\begin{subfigure}{0.45\textwidth}
  \includegraphics[width=\linewidth]{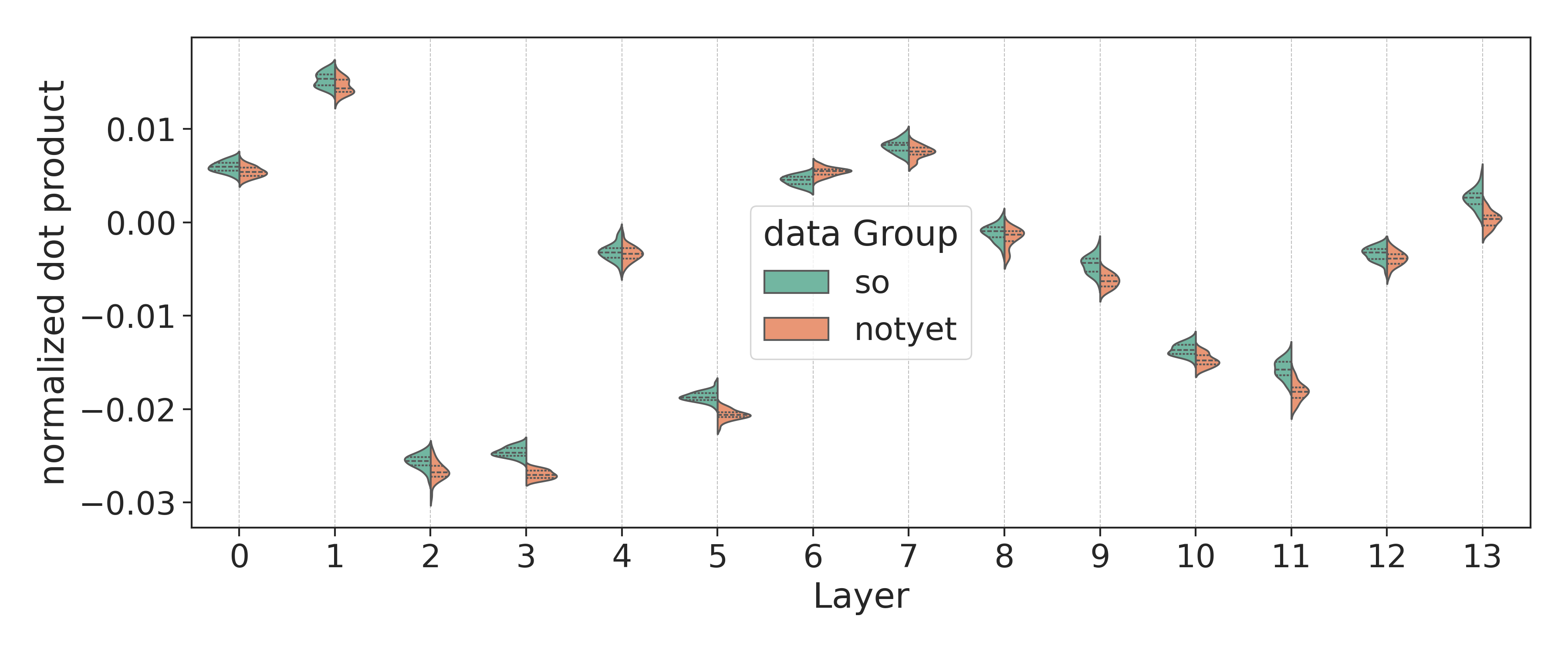}
  \caption{}
  \label{subfig:logatt3}
\end{subfigure}
\begin{subfigure}{0.45\textwidth}
  \includegraphics[width=\linewidth]{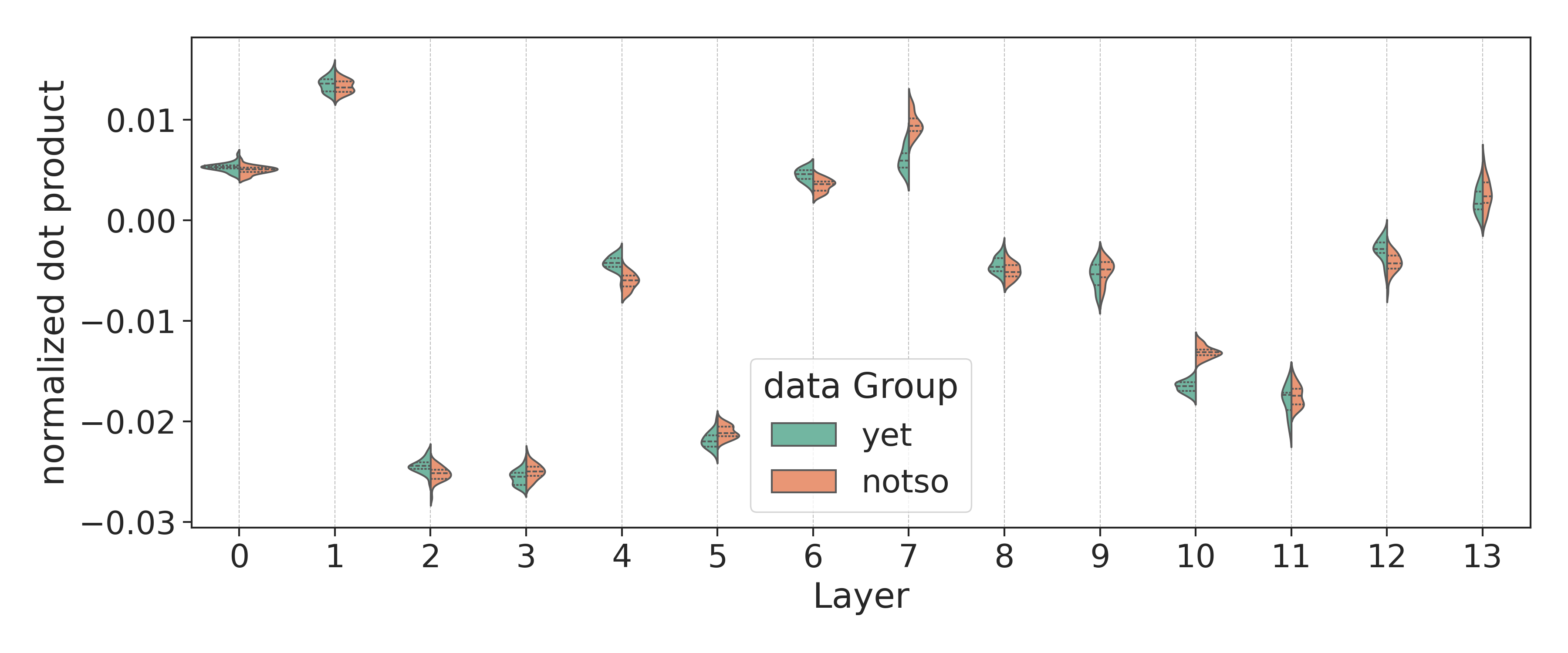}
  \caption{}
  \label{subfig:logatt4}
\end{subfigure}
\caption{Distribution of dot products between layer outputs and logit difference direction. 
(a) layerwise dot product distributions for `able' vs. `unable' answers for positive verbs. 
(b): comparisons across layers 0–13 for different verb types for answer type `able.'
(c): For positive verb type comparisons across layers 0–13 for different combinations with correct answer `able.' This explains the bimodality presented in (b).
(d): As c for negative verbs.}
\label{fig:logatt}
\end{figure*}

Building on these observations, we next identify which layers favor the correct answer. To do this, we 
analyze logit attribution by computing the dot product between each layer's output and the logit direction, where a positive result indicates that a layer's output pushes the model towards the correct answer.
\Cref{subfig:logatt1} presents these attribution results for the attention outputs for positive verbs.
Notably, layers such as 2, 3, 8, 10, and 12 tend to push the output in the negative direction—favoring ``unable'' irrespective of the right answer. 
 In contrast, outputs from attention heads in layers 6, 7, 15, and 22 predominantly drive the prediction 
 towards ``able.''

\Cref{subfig:logatt2} shows dot product values for `able' across positive and negative sentiment verbs.
While the overall magnitude of the dot product remains similar for both verb types, the distributions are sometimes bimodal (e.g., layers 3 and 5 for positive verbs, and layer 10 for negative verbs in \Cref{subfig:logatt2}). To understand the source of the bimodality we further plot for subclasses that answer `able' for positive and negative verbs in  \Cref{subfig:logatt3,subfig:logatt4}.
As discussed in \autoref{ssec:data}, for positive verbs, the answer `able' can be elicited by the discourse marker ``so,'' and by ``yet'' when the verb is negated with ``not'' (covering both causation and antithesis). The plots at layers 3 and 5 in \Cref{subfig:logatt3} clearly show two distinct kernels corresponding to each construction, accounting for the observed bimodality. A similar pattern for negative verbs is illustrated in \autoref{subfig:logatt4}. 

\subsection{RQ3: Which concepts aid model decision-making?}
\begin{figure*}[t]
    \centering 
\begin{subfigure}{0.32\textwidth}
  \includegraphics[width=\linewidth]{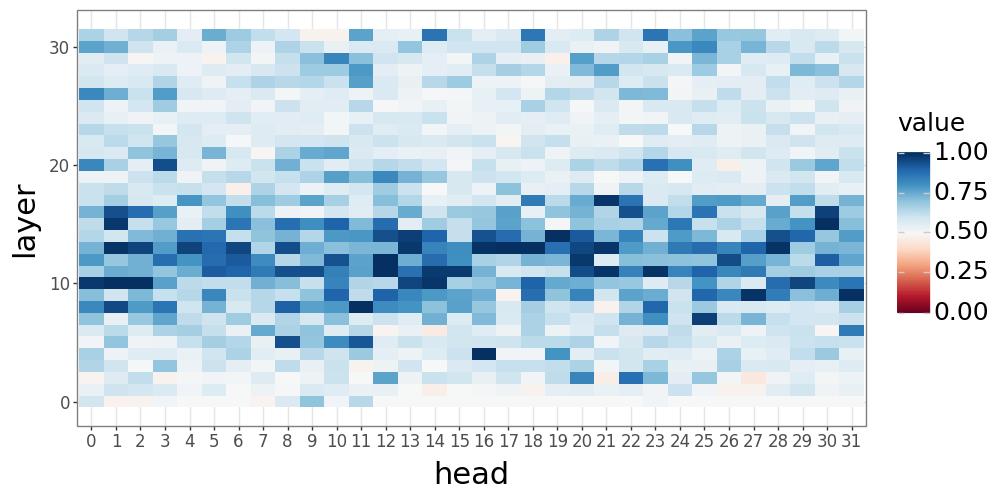}
  \caption{}
  \label{fig:1}
\end{subfigure}\hspace{0.15cm} 
\begin{subfigure}{0.32\textwidth}
  \includegraphics[width=\linewidth]{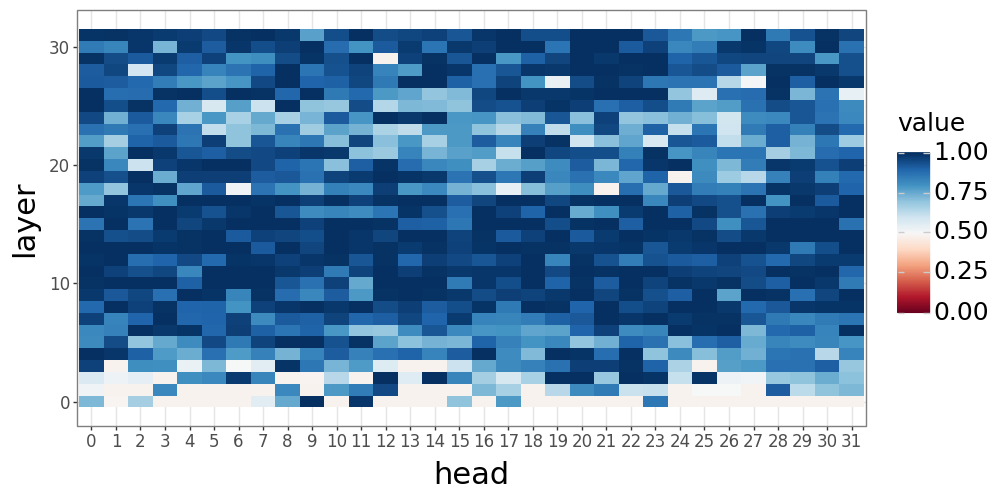}
  \caption{}
  \label{fig:2}
\end{subfigure}\hspace{0.15cm} 
\begin{subfigure}{0.32\textwidth}
  \includegraphics[width=\linewidth]{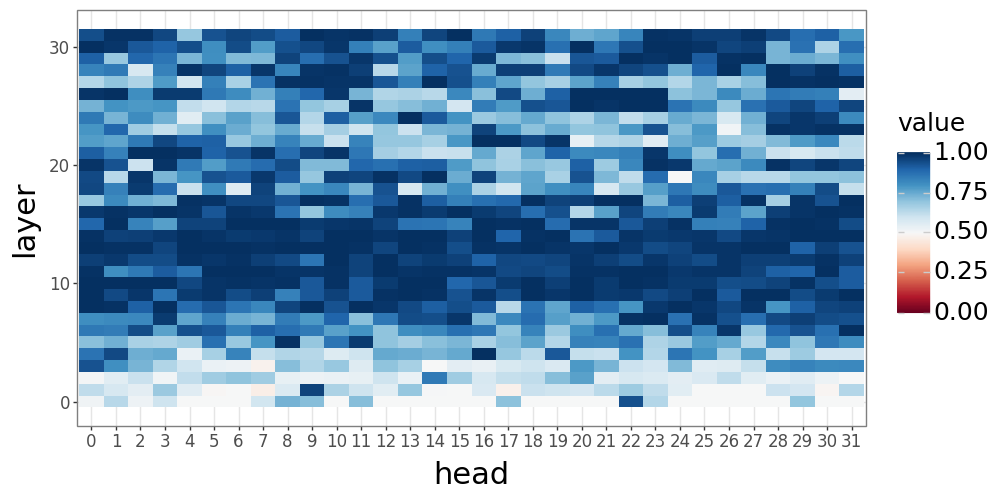}
  \caption{}
  \label{fig:3}
\end{subfigure}

\medskip
\begin{subfigure}{0.32\textwidth}
  \includegraphics[width=\linewidth]{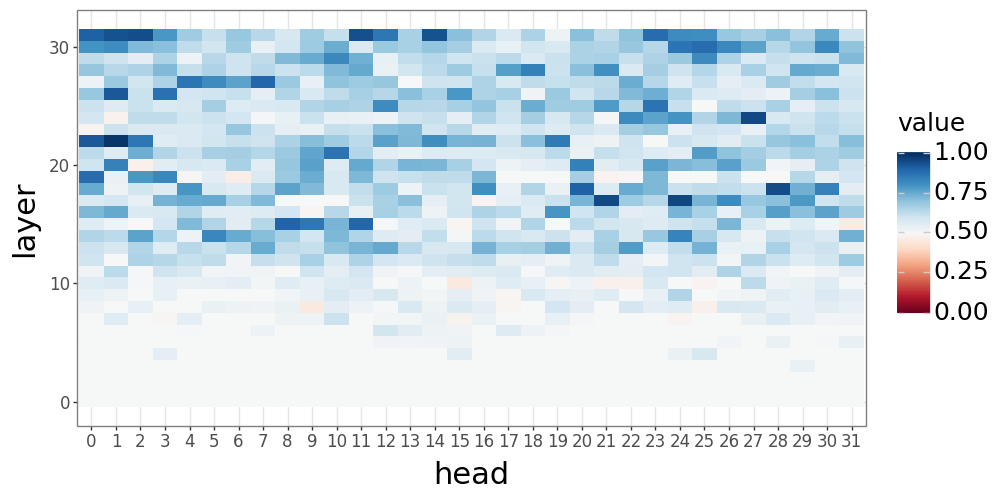}
  \caption{}
  \label{fig:4}
\end{subfigure}\hspace{0.15cm} 
\begin{subfigure}{0.32\textwidth}
  \includegraphics[width=\linewidth]{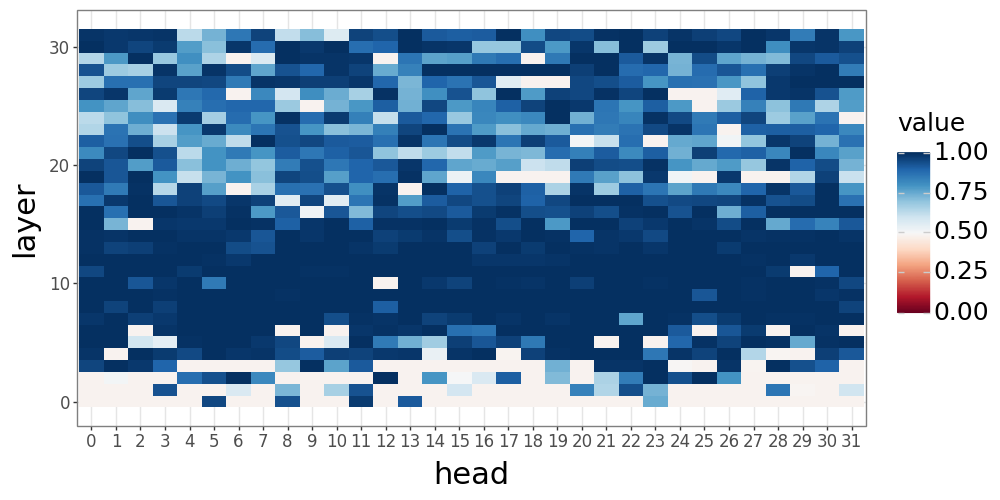}
  \caption{}
  \label{fig:5}
\end{subfigure}\hspace{0.15cm} 
\begin{subfigure}{0.32\textwidth}
  \includegraphics[width=\linewidth]{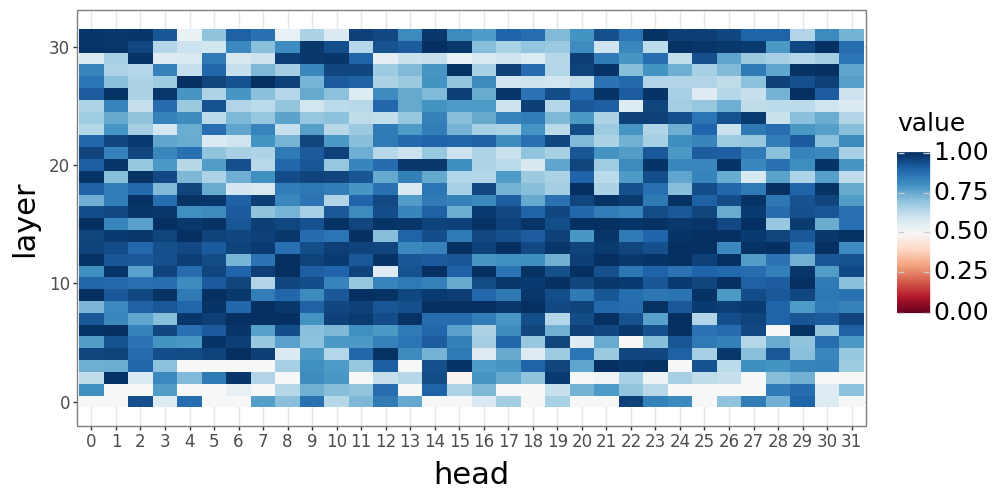}
  \caption{}
  \label{fig:6}
\end{subfigure}
\caption{Accuracy of logistic regression probes across attention heads of LLaMA. The top row shows probe accuracy at the token position for ``so''\slash``yet'', while the bottom row shows accuracy at the position of the last token in the input. Subfigures (a) and (d) show classification accuracy for \emph{``able''}vs. \emph{``unable''}; (b) and (e) show accuracy for classifying the sentiment polarity of the satellite verb; and (c) and (f) show accuracy for predicting the sentiment of the satellite clause.}
\label{fig:logprob}
\end{figure*}

To identify which concepts aid model decision-making and evaluate the potential factors raised in \Cref{ssec:data}, we next analyze the performance of logistic regression probes at each attention head and layer.
 \autoref{fig:logprob} shows probe accuracies for hidden representations at both the ``so''\slash``yet'' and end token 
 positions. We train probes for every head-layer 
 combination for the three classification tasks described in 
 \Cref{ssec:methods}--``able'' versus ``unable,'' verb sentiment in the satellite, and overall sentiment of the satellite--allowing us to pinpoint which internal representations are most predictive and thus likely central to the model's reasoning process.
 
For the ``able'' vs. ``unable'' prediction at the ``so''\slash``yet'' token, the highest probe accuracies 
are found in heads from layers 8–16 (\autoref{fig:logprob}a), with several heads in layers 6–7 also surpassing 0.5 
accuracy. Notably, certain lower-level heads, such as head 12 and 20 in layer 3, achieve near-perfect accuracy. 
Similar patterns are observed in both \emph{sat-verb-sentiment} and \emph{sat-clause-sentiment} classification, where attention heads in the vicinity of layer 10 ($\pm5$) consistently perform well.
These outcomes highlight that the model is actively representing and encoding the high-level semantic cues such as verb sentiment and sentiment of the whole satellite clause.
Additionally, the near-perfect accuracy of some lower-level heads indicates that early 
representations also contribute significantly to the decision process at the ``so''\slash``yet'' location. 
This is further supported by our activation patching analysis (\autoref{subfig:actpatching13}), which demonstrates 
that perturbations at these positions can influence the model's output, confirming the functional importance of these 
encoded semantic concepts.

In contrast, for ``able'' vs. ``unable'' prediction at the final token, distinct performance patterns 
are less evident. Most heads after layer 11 demonstrate high accuracy, with layers 13 and 14 containing several heads 
approaching perfect classification. This aligns with our previous observation that activation patching at these higher 
layers' attention can briefly alter the output (\autoref{fig:logprob}e), suggesting strong decision signals. 
Interestingly, some heads in upper layers exhibit lower activity for classification, implying a potential role in 
supporting or refining the decision process.


\subsection{RQ4: How does information propagate}
To understand how information propagates through the model layers, we employ edge attribution patching (EAP) as described in \autoref{ssec:methods}. To address the issue of vanishing gradients \citep{eap1,eap3}, we also apply integrated gradients (IG) \citep{sundararajan2017,eap3} in conjunction with attribution patching (EAP-IG). Experiments are conducted using both 0-shot and few-shot settings. For each setting, we report the top 1,000 edges identified. Comparing the intersection of edges from both experiments, we find that $51\%$ of the edges are shared, suggesting a stable core of influential connections that consistently support information propagation, regardless of the shot setting.

When analyzing negative verbs, the overlap across shot settings drops to $33.9\%$, indicating greater instability in the set of important edges.
This raises the question of whether the model relies on different pathways to process verbs with negative sentiment compared to verbs with positive sentiment, suggesting that EAP-IG may be more sensitive to shot configuration or verb type in these cases.

Thus, to ascertain whether the model relies on different pathways to process negative verbs, we keep the shot configuration fixed and compare edge overlap across verb types. The relatively high overlaps ($60.8\%$ for 0-shot; $79.4\%$ for 2-shot) indicate that information propagation patterns are more consistent across verb types within the same prompt setup, especially when more context (2-shot) is provided.

\section{Related Work}
    Many techniques have been proposed to uncover a model's inner workings by reverse engineering its computational processes \citep{causal_Geiger,Geiger2023FindingAB,rome_meng,boundlessdas,acdc,eap2,eap1,eap3,zhang2024towards}. These mechanistic interpretability methods have been applied to reveal model behavior in tasks such as indirect object identification, the greater-than task, question answering, and more \citep{ioi,greaterthan,WiegreffeTBHS25}. However, this line of work, while providing insights into model internals, often overlooks the rich linguistic structures underlying text data.

Recently, researchers have begun to explore model understanding with construction grammar \citep{scivetti-etal-2025-unpacking,scivetti-schneider-2025-construction,boguraev-etal-2025-causal,rozner-etal-2025-constructions}, integrating deeper linguistic theory into the analysis of LLMs. In this paper, we advance this line of inquiry by analyzing models from a discourse perspective. Specifically, we examine model behavior through the lens of Rhetorical Structure Theory \citep[RST;][]{rst1,rst2}.
We opt to use RST (as opposed to other discourse modeling frameworks such as PDTB \citep{webber2019penn}) due to its structured yet global approach. Additionally, its nucleus-satellite construction eases delineation discourse scope.

Most related to our work, \citet{miao-kan-2025-discursive} recently investigate factors promoting discourse understanding in Transformers, with a focus on circuit discovery. Similarly to our approach, they use a next-token prediction task with constrained options (e.g., Arg2 or Arg`2) to study how circuits encode discourse relations through ``discursive circuits.'' In contrast, our work takes a broader perspective by applying multiple interpretability techniques and focusing on two key RST relations: causation and antithesis.

RST has been leveraged to enhance the explainability of LLMs by fusing RST into neural networks, for cross-lingual natural language inference  \citep{yuan-etal-2025-cross-document} plus entailment reasoning for science question answering \citep{zhang-etal-2024-empowering}.

\section{Conclusion}
    In this paper, we investigate Transformer encoding of discourse relations, which are critical to characterizing the structure of (and thus to understanding) a document.

We frame this task as a next-token prediction task distinguishing causation and antithesis, and apply a range of causal interventions and probing techniques. Our results indicate that early to mid layers are crucial for decision making at discourse markers such as `so' or `yet,' while higher layers at the final token position are responsible for making and maintaining the model's ultimate prediction.
At the circuit level, we observe a stable set of connections that reliably contribute to computation, though additional computational edges may be contextually activated or deactivated depending on the prompt and verb type. Furthermore, we find that certain layers exhibit inherent biases toward specific outcomes. Our results provide insight into how discourse modeling is promoted in language models, giving way to broader model safety and control.
Further, the statistically significant and substantive native of our findings suggests that our methodology is extensible to additional discourse relations and model architectures.

\section{Limitations}
    Like prior work \citep{miao-kan-2025-discursive}, we constrain our task to a next-token prediction task with one of two expected outputs and focus on two RST relations. We opt to study model behavior through the lens of RST. More specifically our analysis is on distinguishing causation and antithesis, formulated as a next-token prediction task. Future work may adapt our approach to examine additional RST relations.

We focus our analyses on Transformer models due to their prevalence in contemporary NLP literature, and in order to promote generalizability we examine both a LLaMA and Mistral model.

This study focuses exclusively on discourse relations in English. 
As discourse structure and linguistic cues can vary substantially across languages, generalization of our results across languages is unknown. 
Furthermore, the observed results may depend on the specific models and discourse relations examined in this work.


\bibliography{custom}

\appendix 

\section{Appendix}\label{sec:appendix}
 \label{appendix_a}
 \begin{figure}[ht]
        \centering
        \begin{tcolorbox}[
            colback=gray!10,      
            colframe=black!50,    
            coltitle=black,            
            boxrule=0.5pt,
        ]
        \textbf{System Prompt: } You are an AI language model trained to choose between the words ``able'' and ``unable'' to complete a partially written sentence. Your decision should be based on the meaning of the sentence and its implied relationship, such as causation or contrast.

        Carefully review the provided examples. Then, for each new sentence, output only the correct word: ``able'' or ``unable.''

        Do not provide explanations. 
        Do not output anything else.\\
        
        \textbf{User Prompt: } 
        1. Sentence: John got the train this morning, so he was \_\_\_ to reach on time.\\
        Answer: able\\
        2. Sentence: The team practiced all week, so they were \_\_\_ to perform well.\\
        Answer: able\\
        3. Sentence: The team practiced all week, yet they were \_\_\_ to perform well.\\
        Answer: unable\\
        4. Sentence: Lena studied hard, yet she was \_\_\_ to pass the exam.\\
        Answer: unable\\

        Now complete the following:\\
        5. Sentence: He slept well last night, yet he was \_\_\_ to run this morning.\\
        Answer:
        \end{tcolorbox}
        \caption{Example with system prompt and few-shot demonstrations.}
        \label{fig:sys_prompt}
    \end{figure}
\begin{figure*}[htb]
    \centering 
\begin{subfigure}{0.3\textwidth}
  \includegraphics[width=\linewidth]{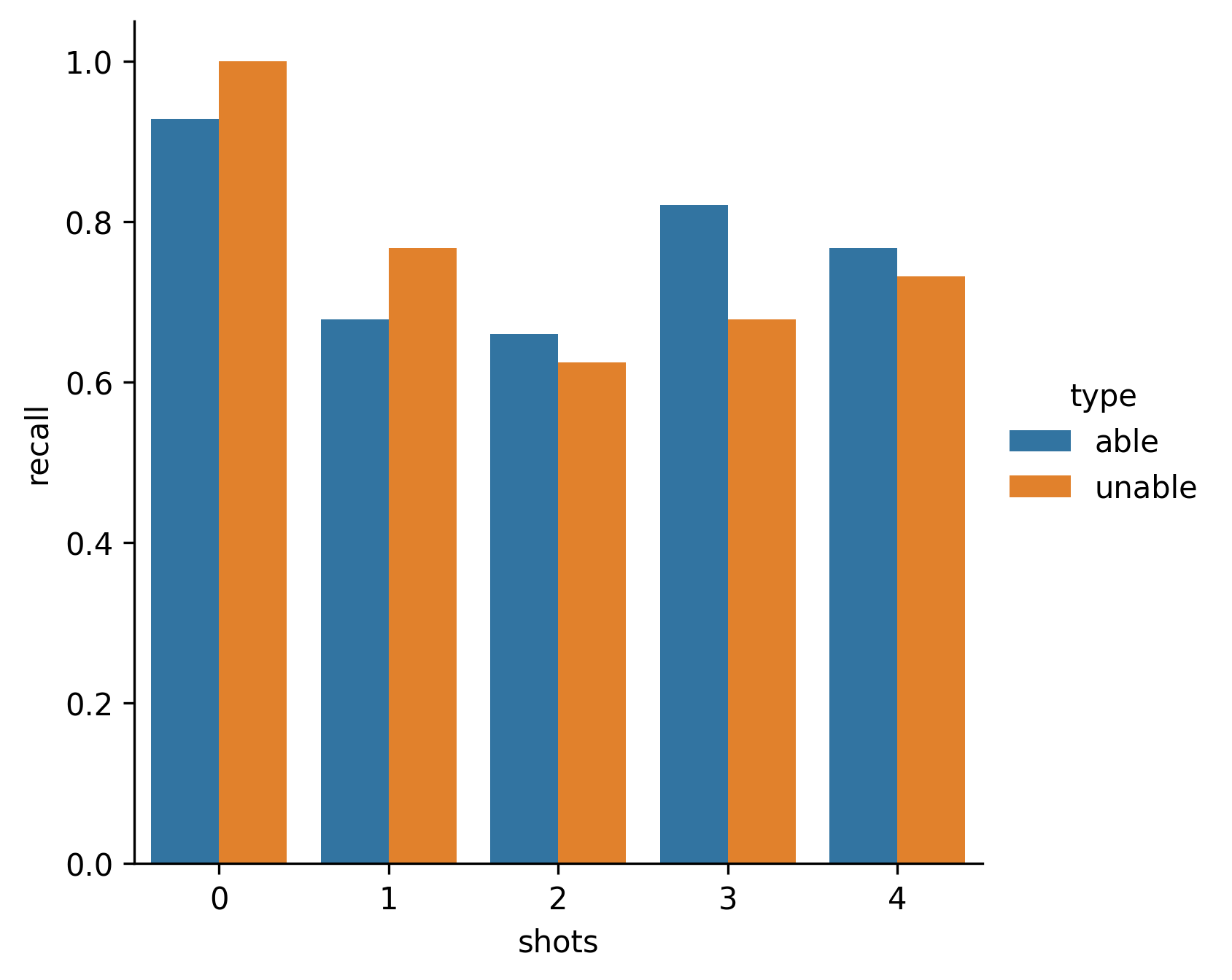}
  \caption{}
  \label{subfig:llama_overall_recall}
\end{subfigure}\hspace{0.15cm} 
\begin{subfigure}{0.30\textwidth}
  \includegraphics[width=\linewidth]{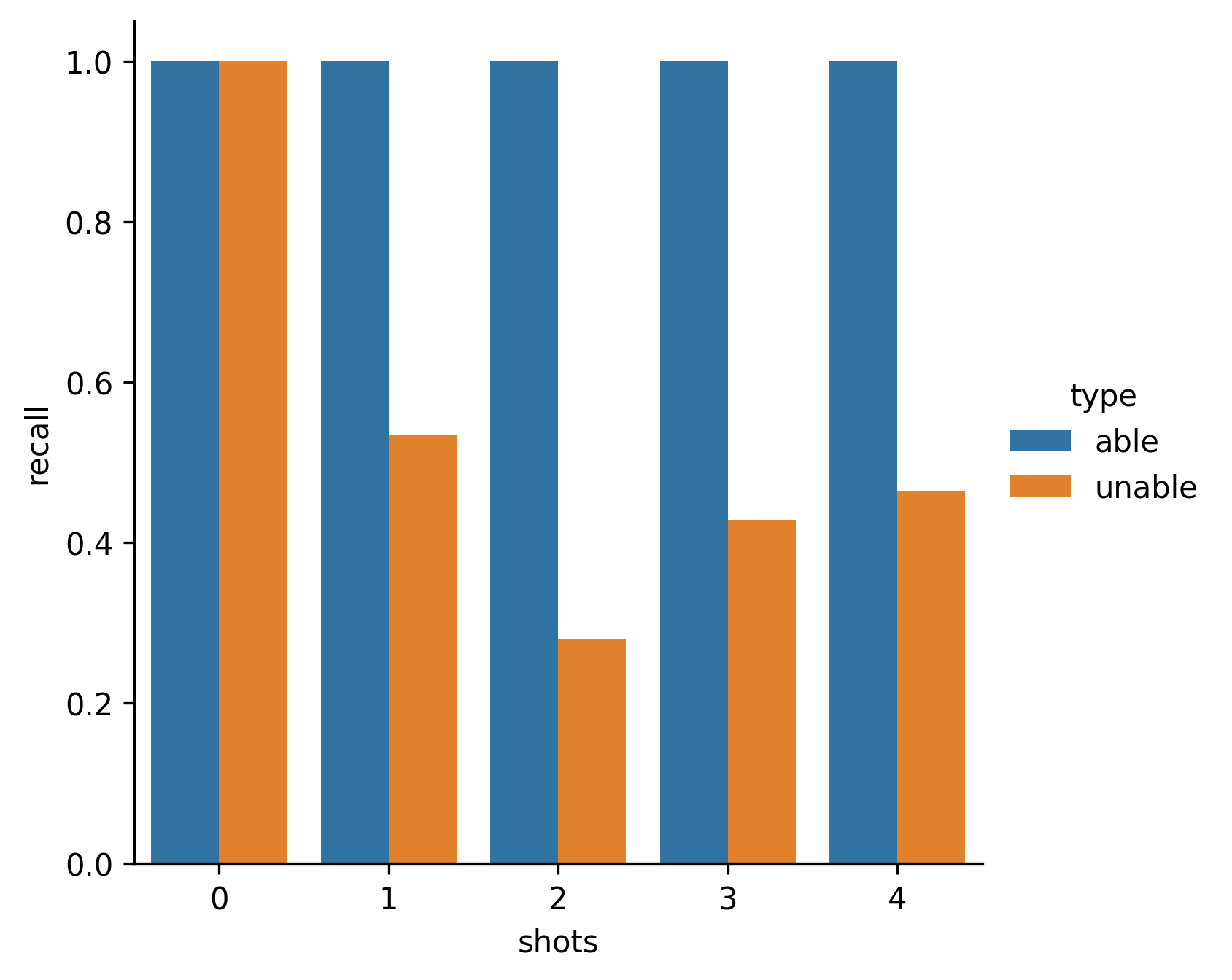}
  \caption{}
  \label{subfig:llama_wonot_recall}
\end{subfigure}\hspace{0.15cm} 
\begin{subfigure}{0.3\textwidth}
  \includegraphics[width=\linewidth]{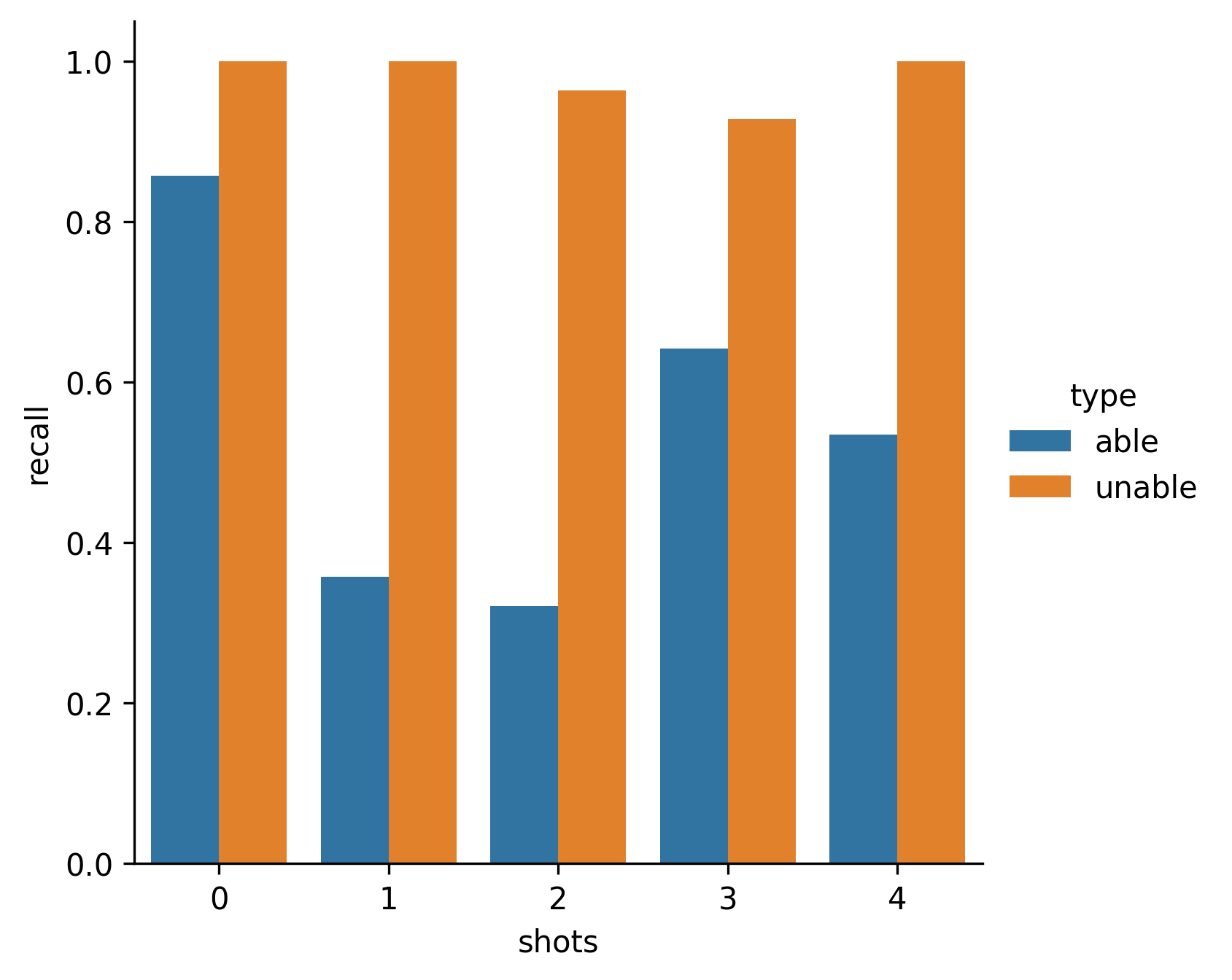}
  \caption{}
  \label{subfig:llama_wnot_recall}
\end{subfigure}
\medskip
\begin{subfigure}{0.3\textwidth}
  \includegraphics[width=\linewidth]{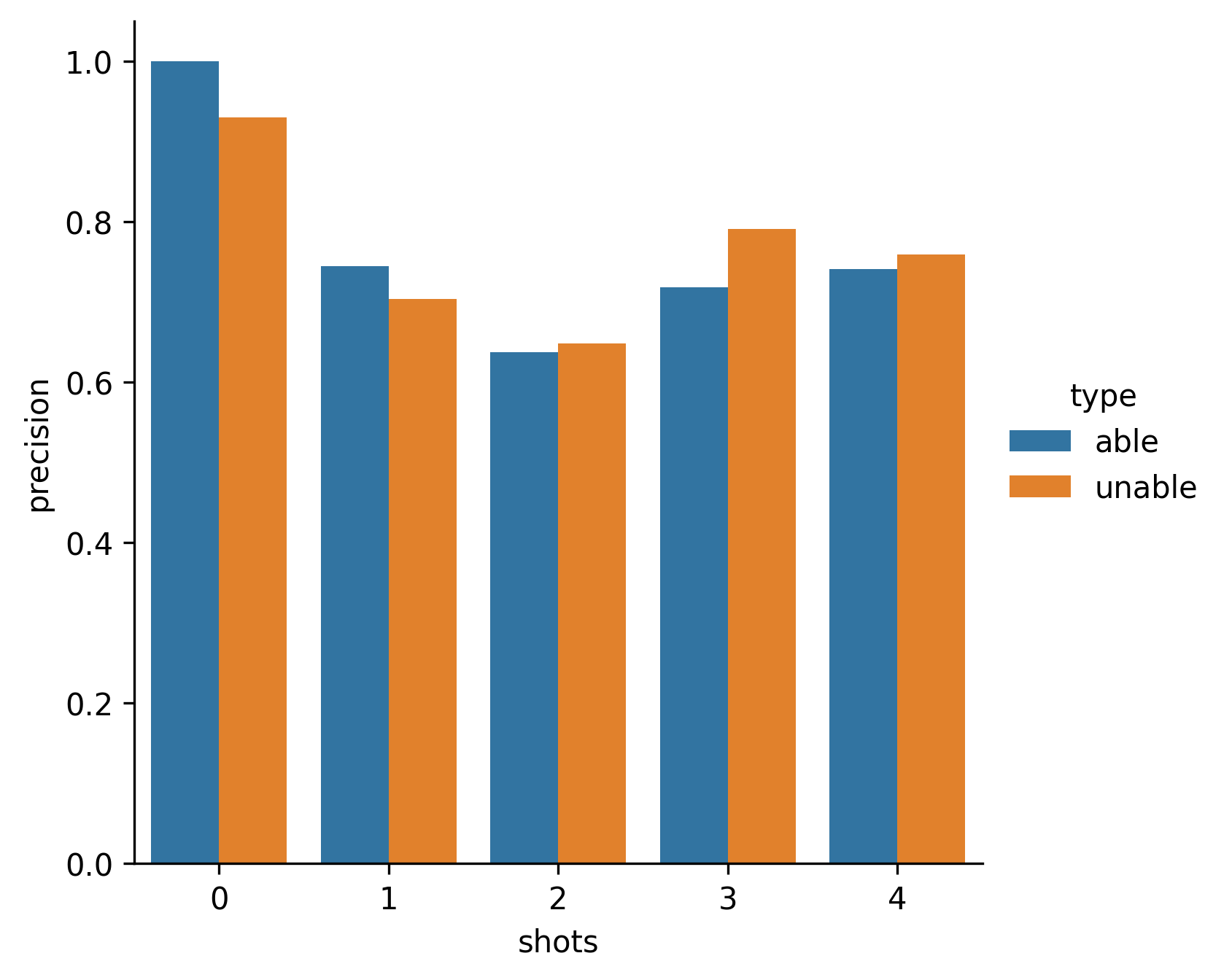}
  \caption{}
  \label{subfig:llama_overall_precision}
\end{subfigure}\hspace{0.15cm} 
\begin{subfigure}{0.30\textwidth}
  \includegraphics[width=\linewidth]{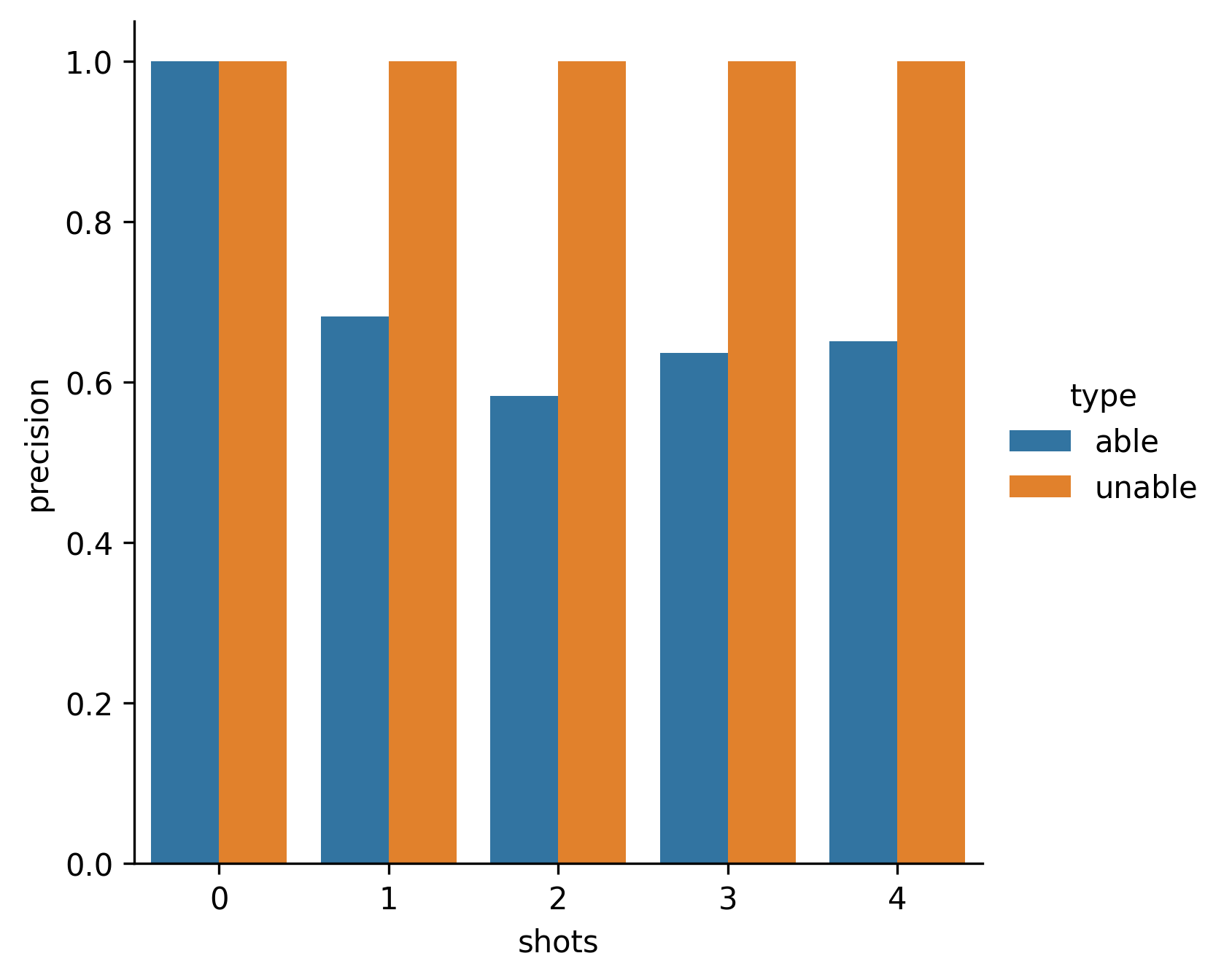}
  \caption{}
  \label{subfig:llama_wonot_precision}
\end{subfigure}\hspace{0.15cm} 
\begin{subfigure}{0.3\textwidth}
  \includegraphics[width=\linewidth]{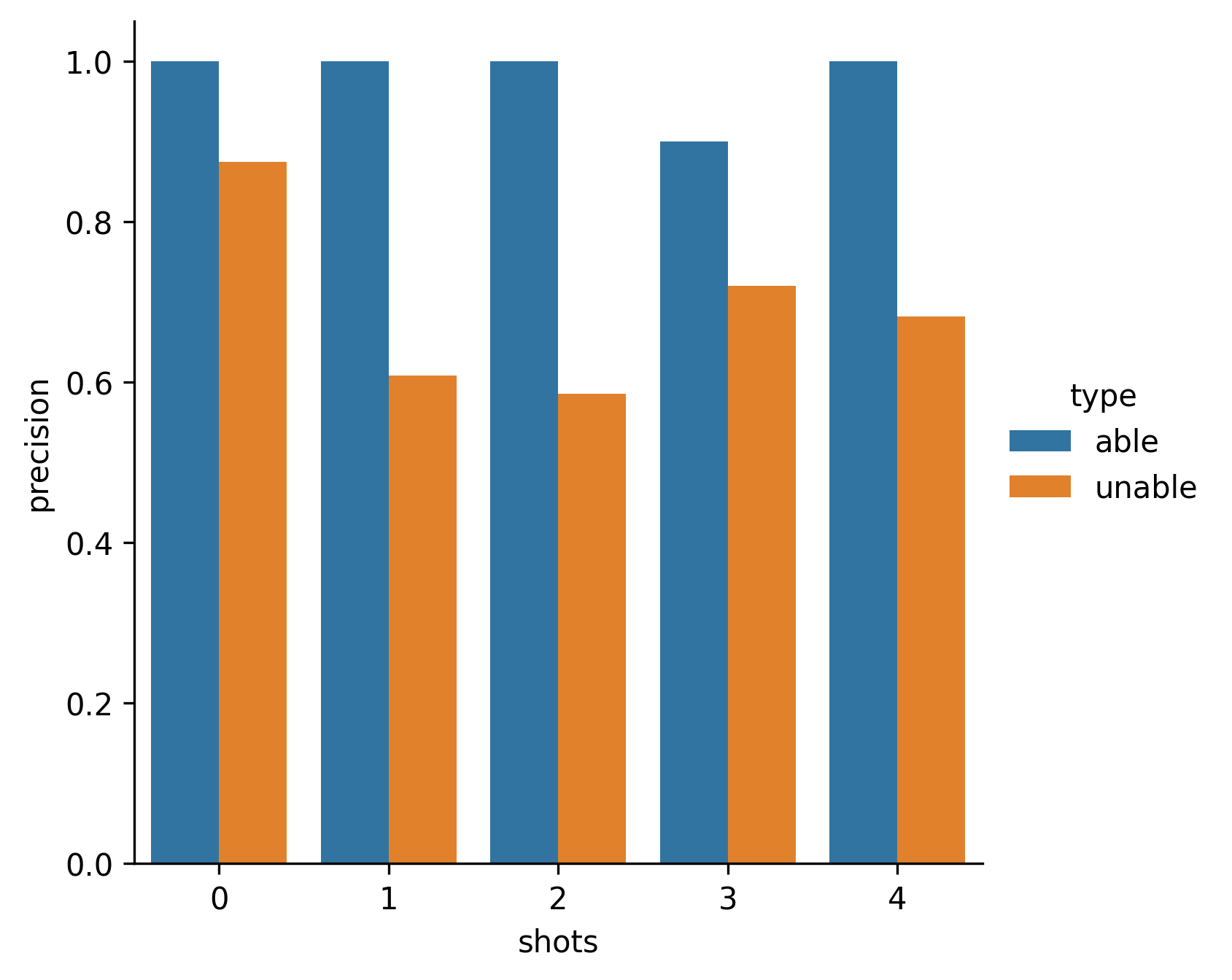}
  \caption{}
  \label{subfig:llama_wnot_precision}
\end{subfigure}
\caption{LLaMA performance on examples with positive verbs. The top row shows recall, and the bottom row shows precision. (a) and (d): overall performance; (b) and (e): without explicit negation; (c) and (f): with negation. }
\label{fig:llama_shots}
\end{figure*}

\paragraph{Number of demonstrations per prompt.} For prompting the model to predict the last token we replace the actual `able' or 'unable' token from the query sentence and present the query sentence along with some example sentences. An example with the system prompt is demonstrated in \Cref{fig:sys_prompt}.
We use two demonstrations per sentence type—`so,' `yet,' `NOT-so,' and `NOT-yet.' Following \cite{WiegreffeTBHS25}, we patch only those examples for which the model predicts accurately. To understand the effect of the number of demonstrations in the prompt, we experimented with up to four examples per class. \autoref{fig:llama_shots} shows the comparative recall and precision for next-token predictions of `able' vs. `unable' with the LLaMA model. Interestingly, the model performs better without any demonstrations: for instances without negation, recall is consistently high for `able,' and precision is high for `unable.' This pattern reverses when explicit negation (`not') is introduced in the satellite verb.

We also observe that the order of demonstrations in the prompt impacts performance. Consistency between the query and the provided demonstrations yields perfect recall and precision. For example, queries without `NOT' benefit from demonstrations without `NOT,' and queries with `NOT' benefit from demonstrations with `NOT.' In this setup, we use two shots per class. While the organization, order, and number of demonstrations could be systematically studied as a separate problem, we leave this for future work.


\begin{table*}[htb]
\centering
  \scriptsize
  \begin{tabular}{l|cccccccc}
  \toprule
    & pos-so & pos-yet & negated-pos-so & negated-pos-yet & neg-so & neg-yet & negated-neg-so & negated-neg-yet \\
  \midrule
    pos-so       & - & diff & diff & diff & diff & diff&diff&diff \\
    pos-yet      & diff & - & diff & not diff &diff &not diff&diff&diff\\
    negated-pos-so    & diff & diff & - & diff & diff & diff & not diff & diff \\
    negated-pos-yet &diff &not diff & diff &- & diff & not diff & diff &diff\\
    \hline
    neg-so &diff & diff & diff & diff&- &diff & diff & not diff\\
    neg-yet & diff& not diff & diff & not diff & diff &- & diff & diff \\
    negated-neg-so &diff &diff & not diff&diff &diff & diff&- & diff\\
    negated-neg-yet &diff & diff& diff & diff& not diff & diff & diff&-\\
  \bottomrule
  \end{tabular}
  \caption{Pairwise t-test of normalized SHAP values across constructions at verb position. `pos' is for verb with positive sentiment. `neg' is for verb with negative sentiment. `pos-so' refers to sentences with positive verb and so as markers. `negated-pos-so' refers to example with explicit negation of positive verb and `so' as discourse marker. }
  \label{tab:ttestVerb}
\end{table*}

\begin{figure*}[htb]
    \centering 
\begin{subfigure}{0.3\textwidth}
  \includegraphics[width=\linewidth]{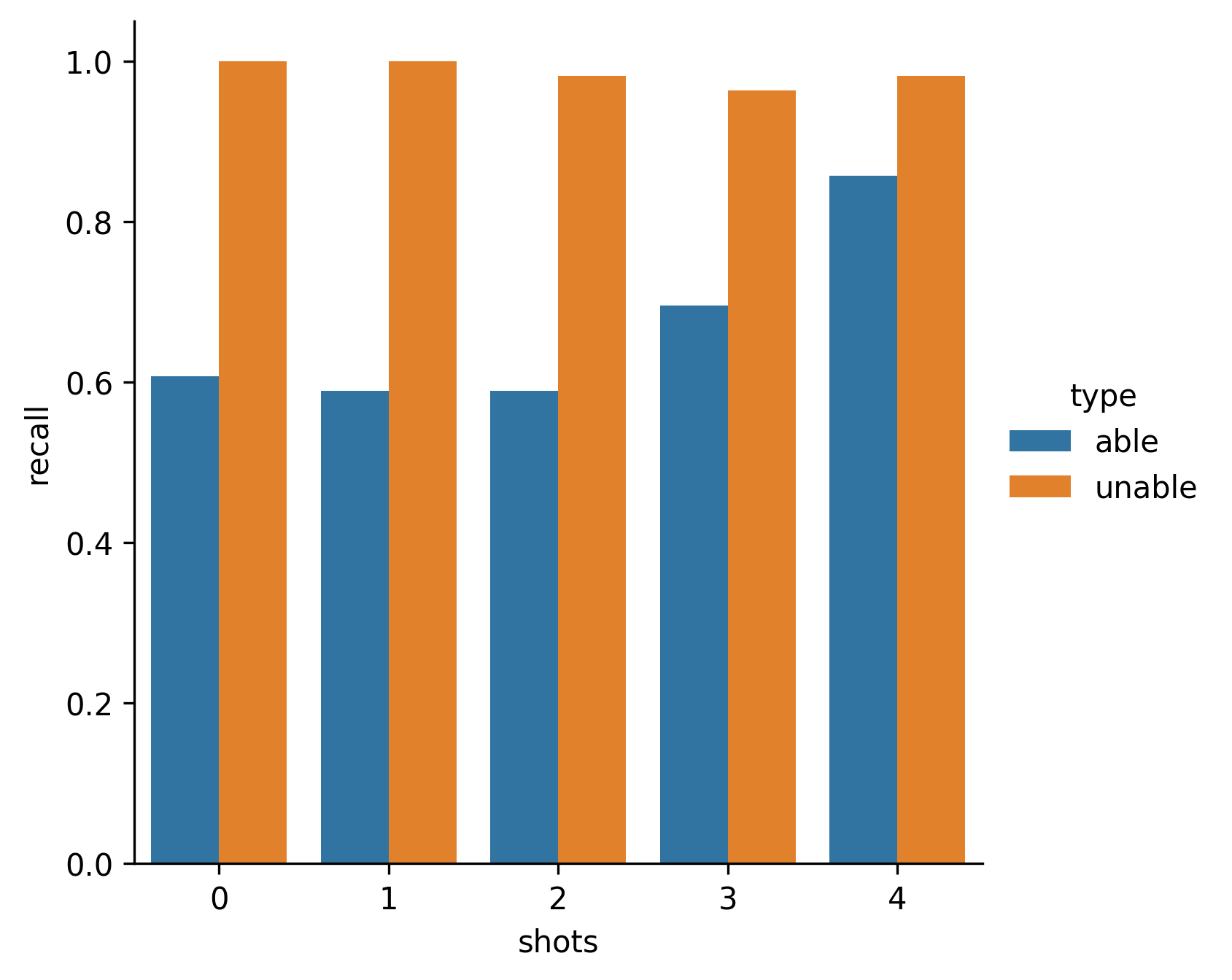}
  \caption{}
  \label{subfig:mistral_overall_recall}
\end{subfigure}\hspace{0.15cm} 
\begin{subfigure}{0.30\textwidth}
  \includegraphics[width=\linewidth]{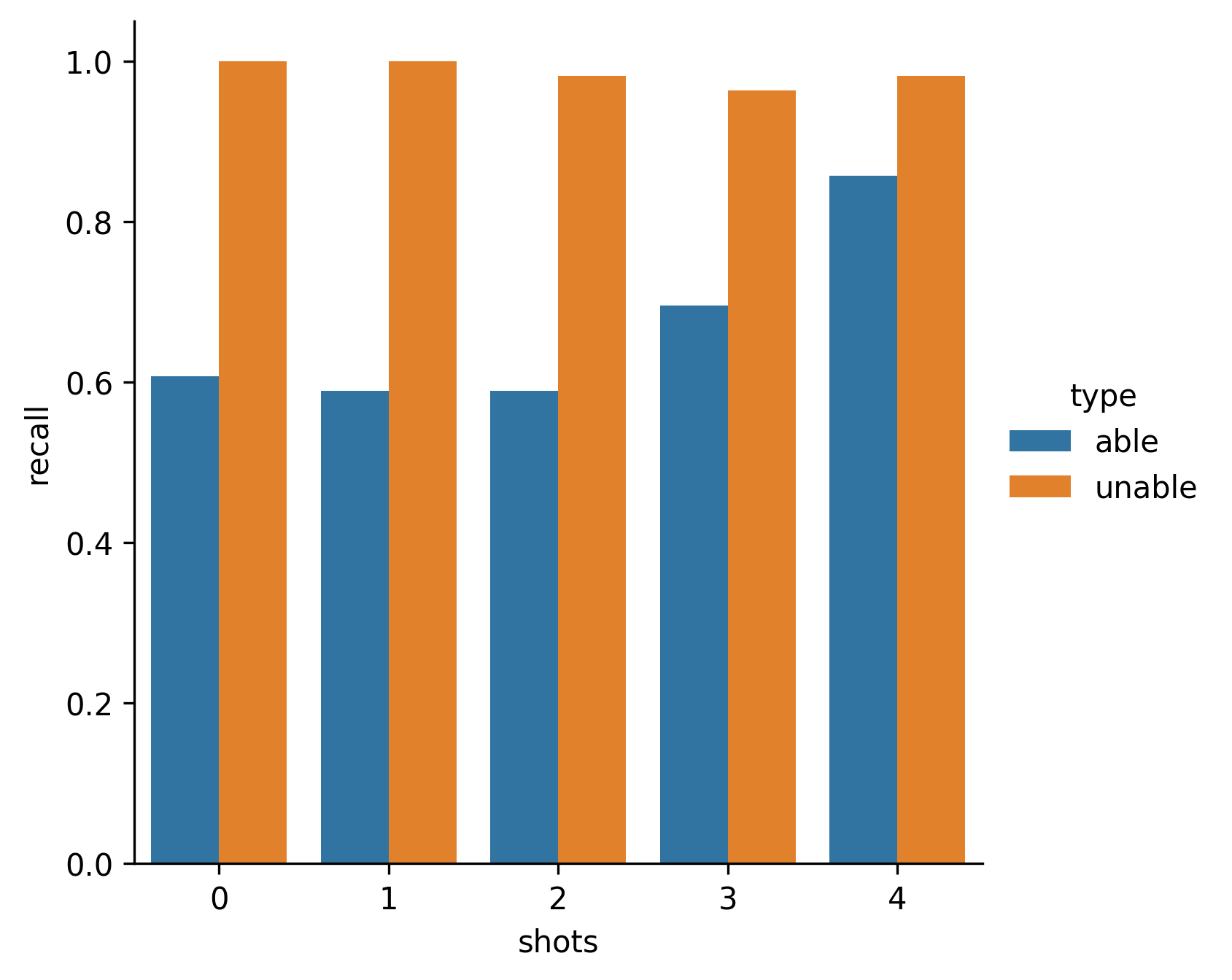}
  \caption{}
  \label{subfig:mistral_wonot_recall}
\end{subfigure}\hspace{0.15cm} 
\begin{subfigure}{0.3\textwidth}
  \includegraphics[width=\linewidth]{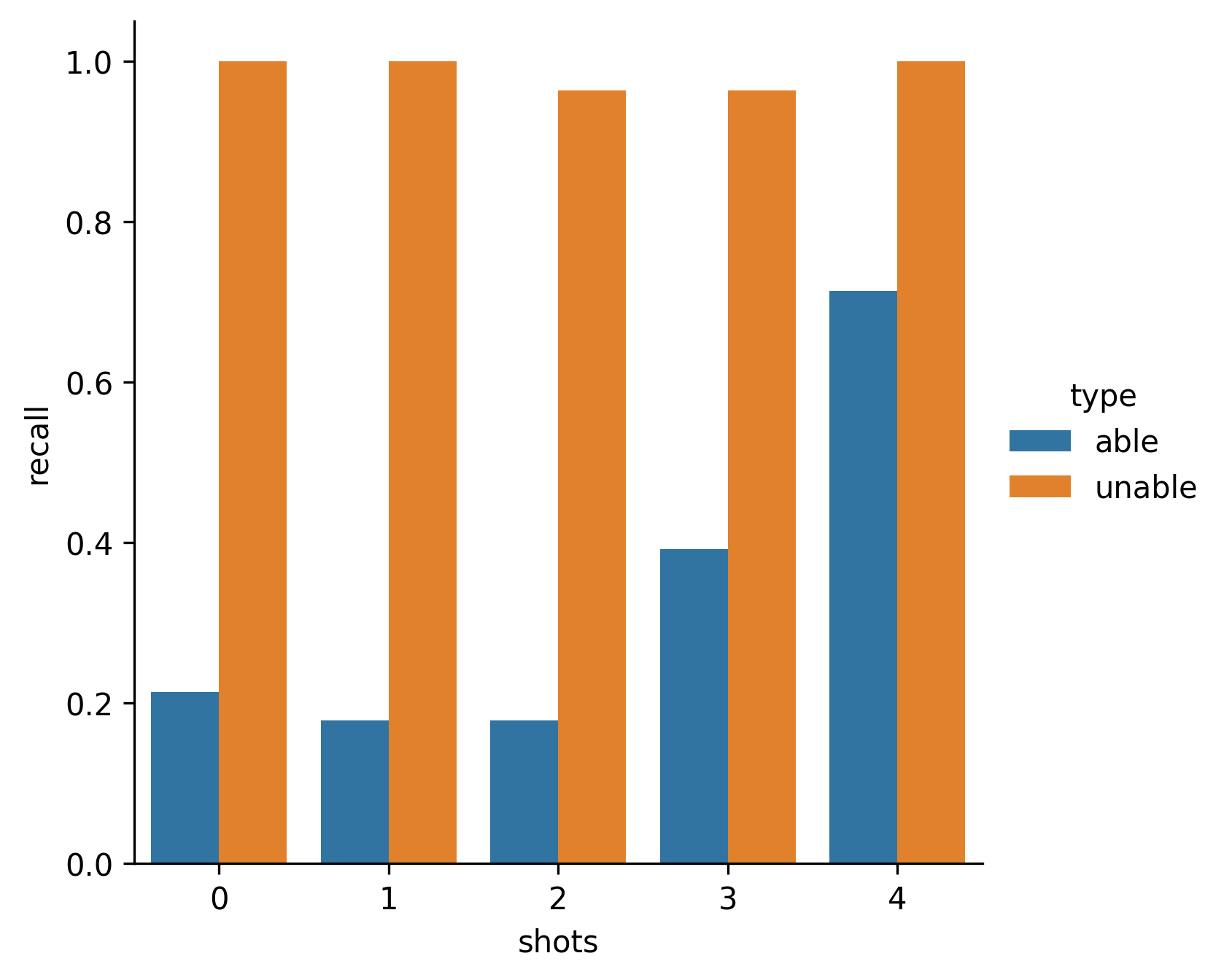}
  \caption{}
  \label{subfig:mistral_wnot_recall}
\end{subfigure}
\medskip
\begin{subfigure}{0.3\textwidth}
  \includegraphics[width=\linewidth]{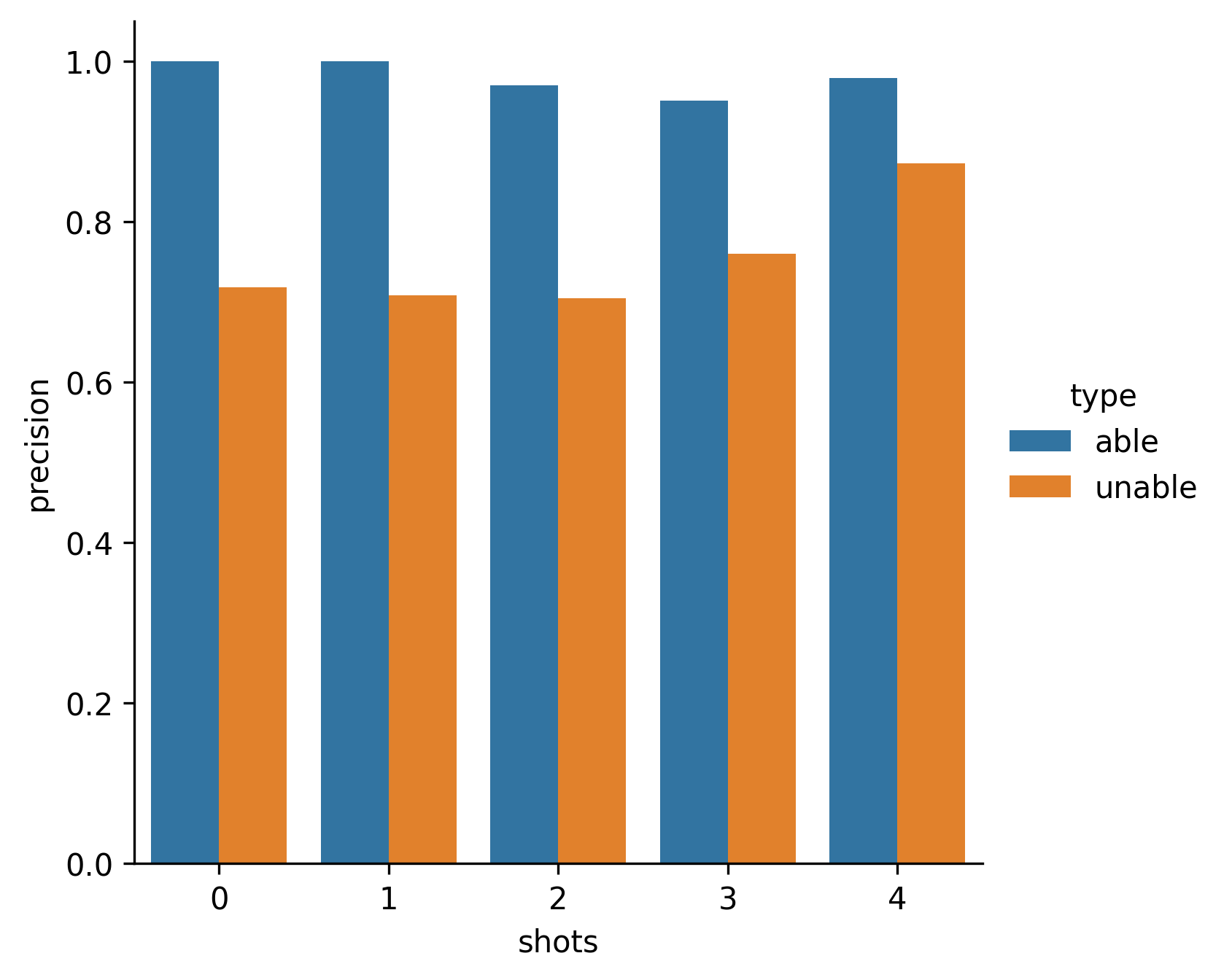}
  \caption{}
  \label{subfig:mistral_overall_precision}
\end{subfigure}\hspace{0.15cm} 
\begin{subfigure}{0.30\textwidth}
  \includegraphics[width=\linewidth]{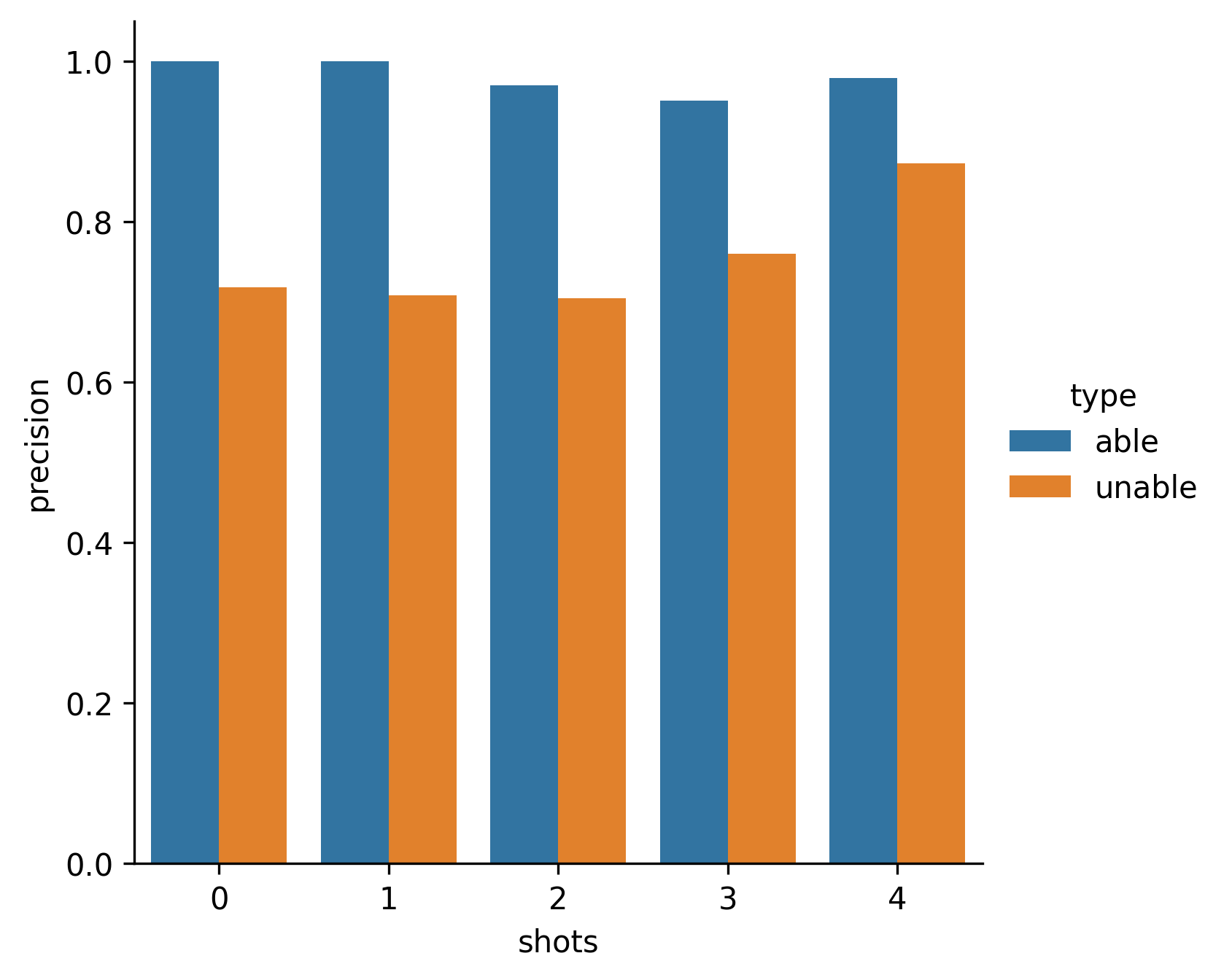}
  \caption{}
  \label{subfig:mistral_wonot_precision}
\end{subfigure}\hspace{0.15cm} 
\begin{subfigure}{0.3\textwidth}
  \includegraphics[width=\linewidth]{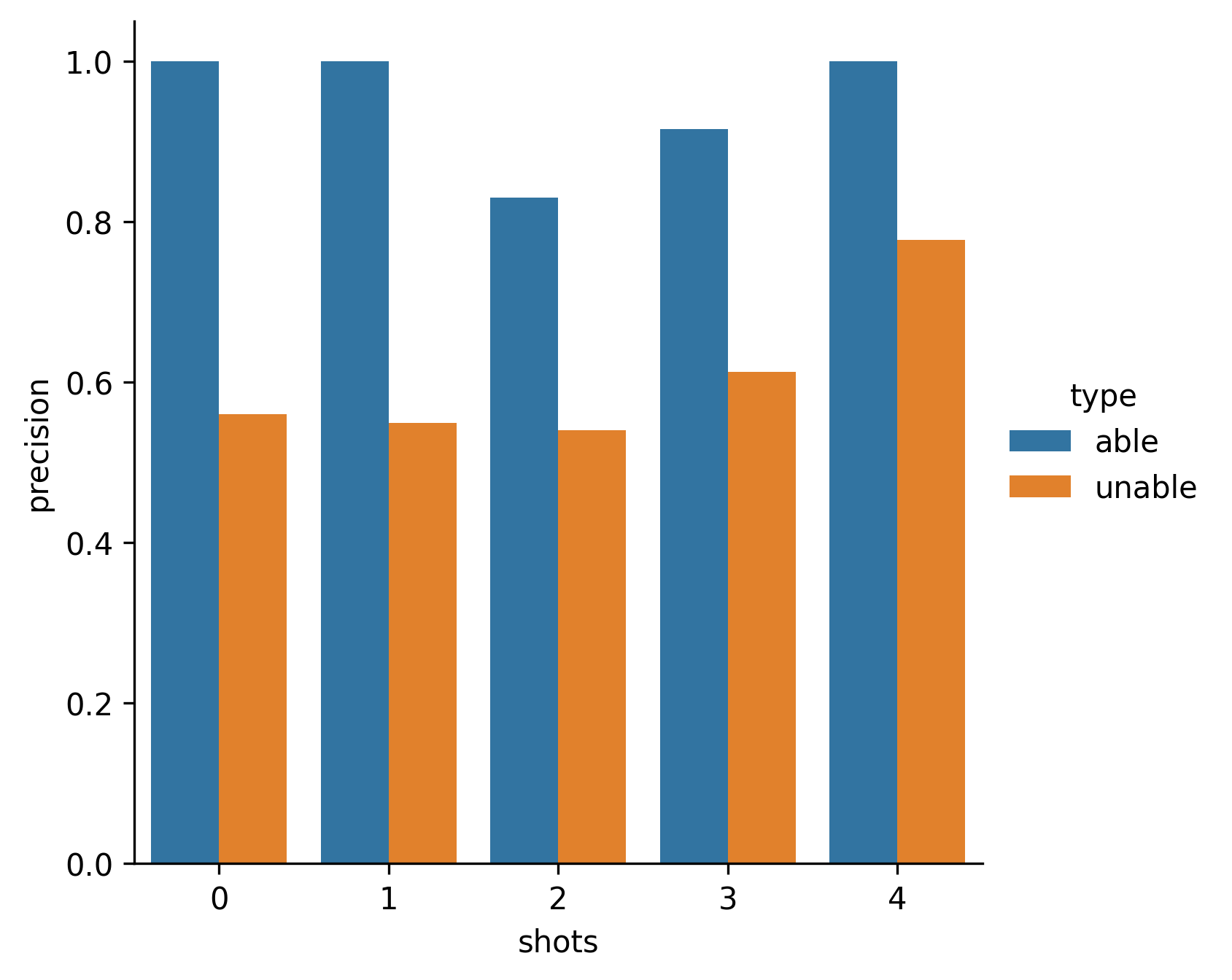}
  \caption{}
  \label{subfig:mistral_wnot_precision}
\end{subfigure}
\caption{Mistral performance on examples with positive verbs. The top row shows recall, and the bottom row shows precision. (a) and (d): overall performance; (b) and (e): without explicit negation; (c) and (f): with negation. }
\label{fig:mistral_shots}
\end{figure*}

\begin{figure*}[htb]
    \centering 
\begin{subfigure}{0.36\textwidth}
  \includegraphics[width=\linewidth]{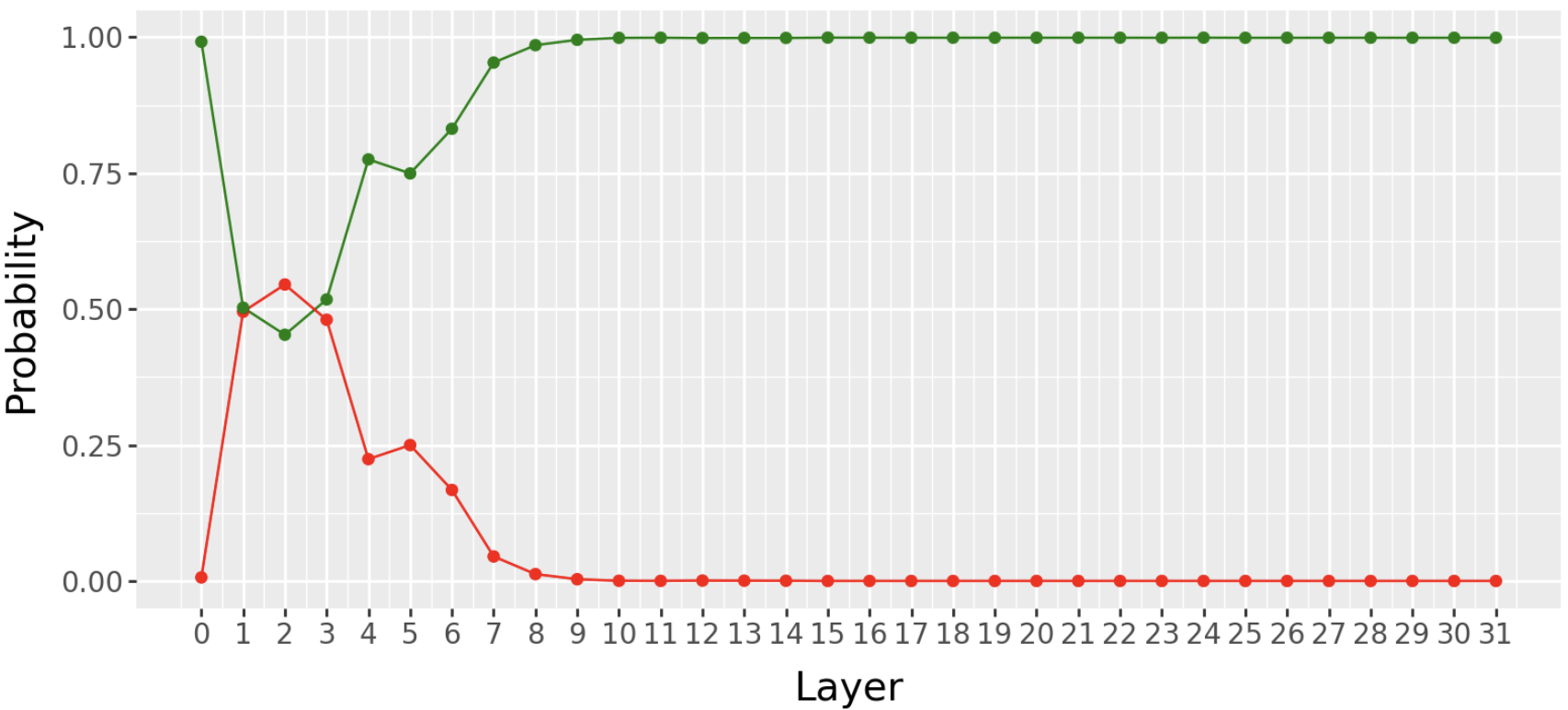}
  \caption{}
  \label{subfig:actpatching11}
\end{subfigure}\hspace{0.15cm} 
\begin{subfigure}{0.46\textwidth}
  \includegraphics[width=\linewidth]{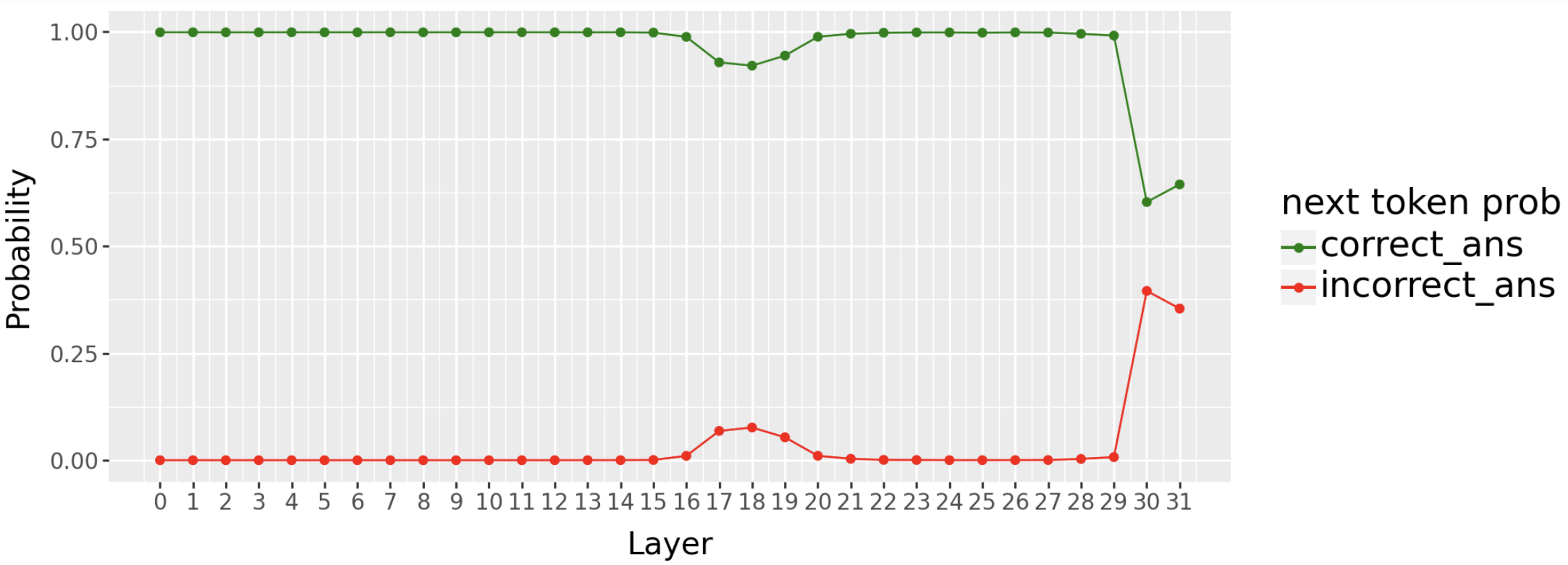}
  \caption{}
  \label{subfig:actpatching14}
\end{subfigure} 
\caption{Change in probability from the correct answer (``able'' or ``unable'') to the wrong answer under activation patching between causation and antithesis with a positive verb.
The left column shows the change in probability when patched at the token location for ``so'' and ``yet,'' while the right column shows the same change at the last token location. (a) and (b) illustrate the effect of patching at the MLP activation. }
\label{fig:mlp_patching}
\end{figure*}

\begin{figure*}[htb]
    \includegraphics[width=\linewidth]{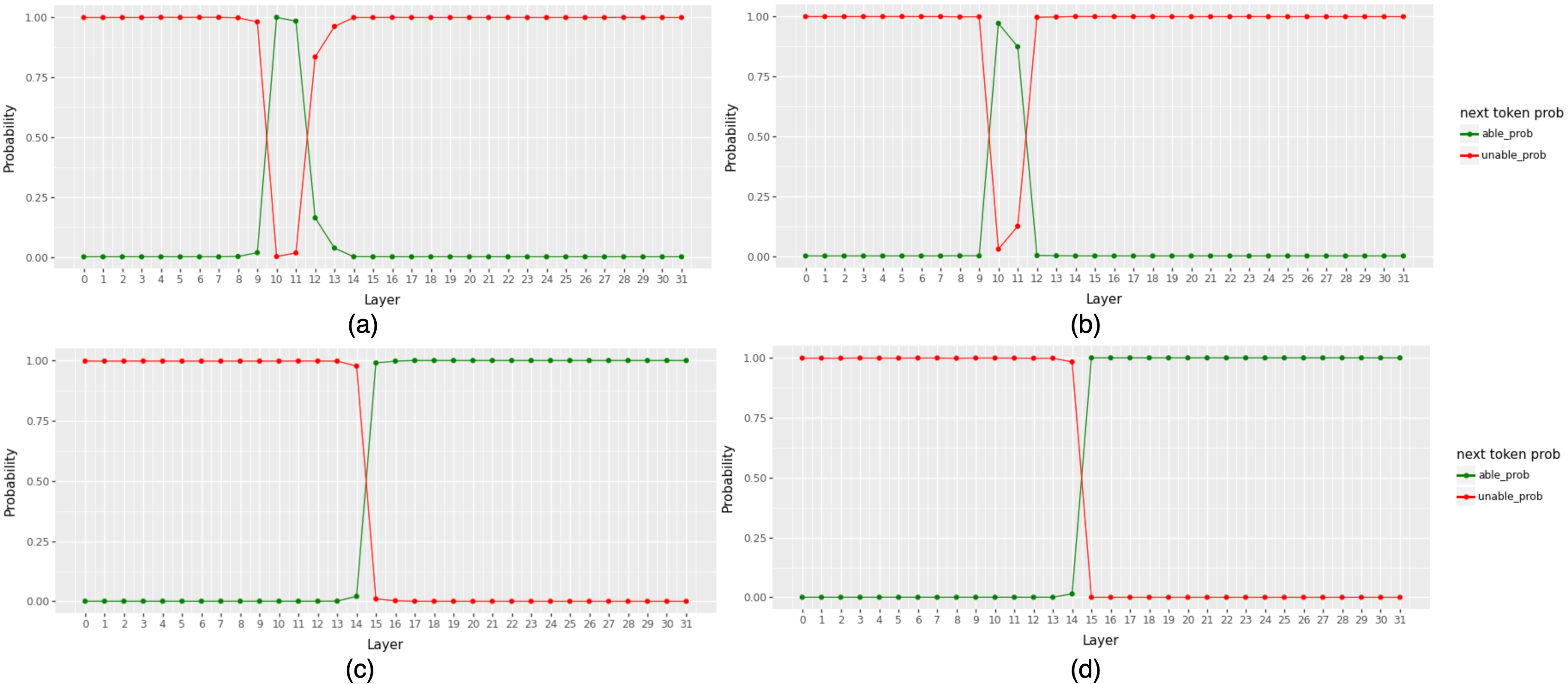}
    \caption{Activation patching in LLaMA with sentence with explicit `NOT' and 'unable' is the correct answer. Top rows depicts when patched at ``so''\slash``yet'' location. Bottom row shows when patched at last token. (a) and (c)patching from source combination not-yet(able) to base combination yet(unable) (b) and (d) patching from source combination so(able) to base combination not-so(unable)}
    \label{fig:patching_with_NOT_Ans_unable}
\end{figure*}

\begin{figure*}[htb]
    \includegraphics[width=\linewidth]{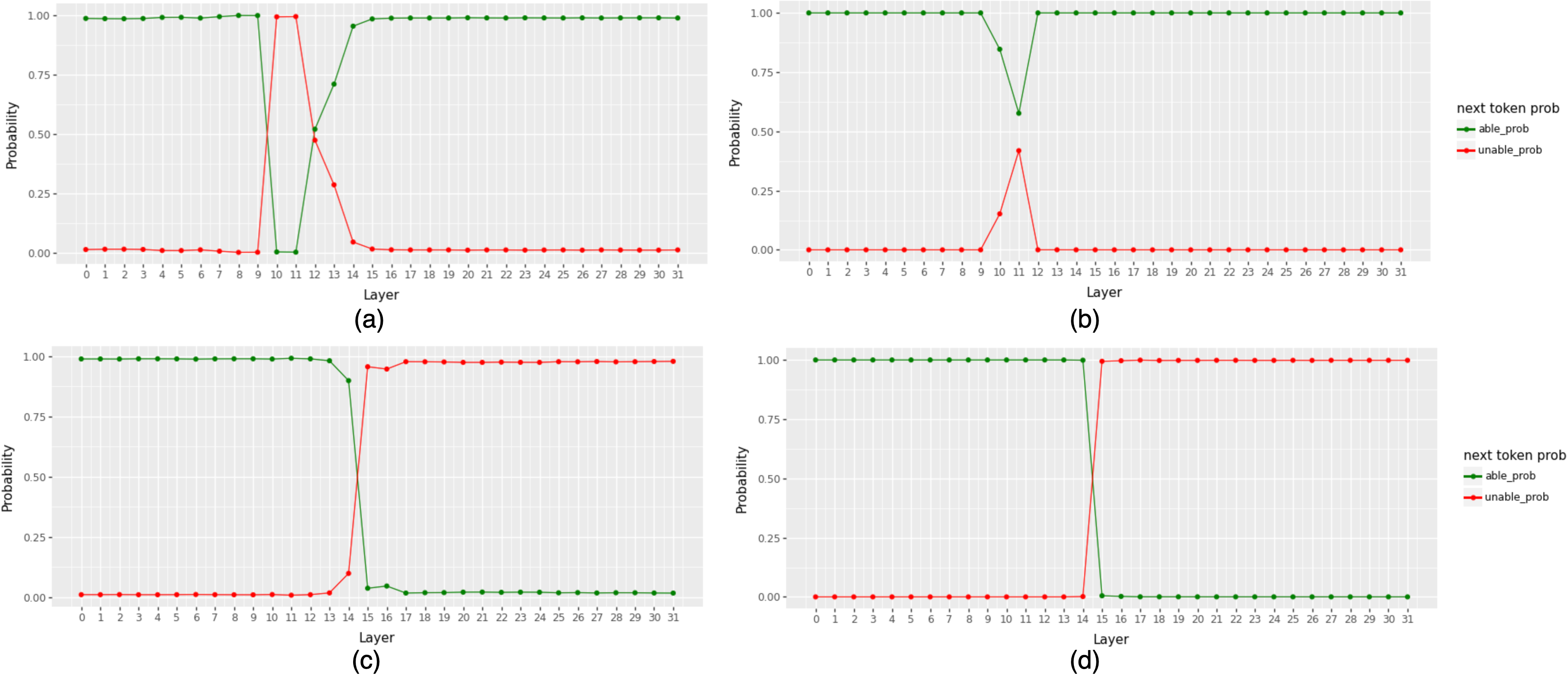}
    \caption{Activation patching in LLaMA with sentence with explicit `NOT' and 'able' is the correct answer. Top rows depicts when patched at ``so''\slash``yet'' location. Bottom row shows when patched at last token. (a) and (c)patching from source combination not-yet(able) to base combination yet(unable) (b) and (d) patching from source combination so(able) to base combination not-so(unable)}
    \label{fig:patching_with_NOT_Ans_able}
\end{figure*}

\begin{figure*}[htb]
    \centering 
\begin{subfigure}{0.31\textwidth}
  \includegraphics[width=\linewidth]{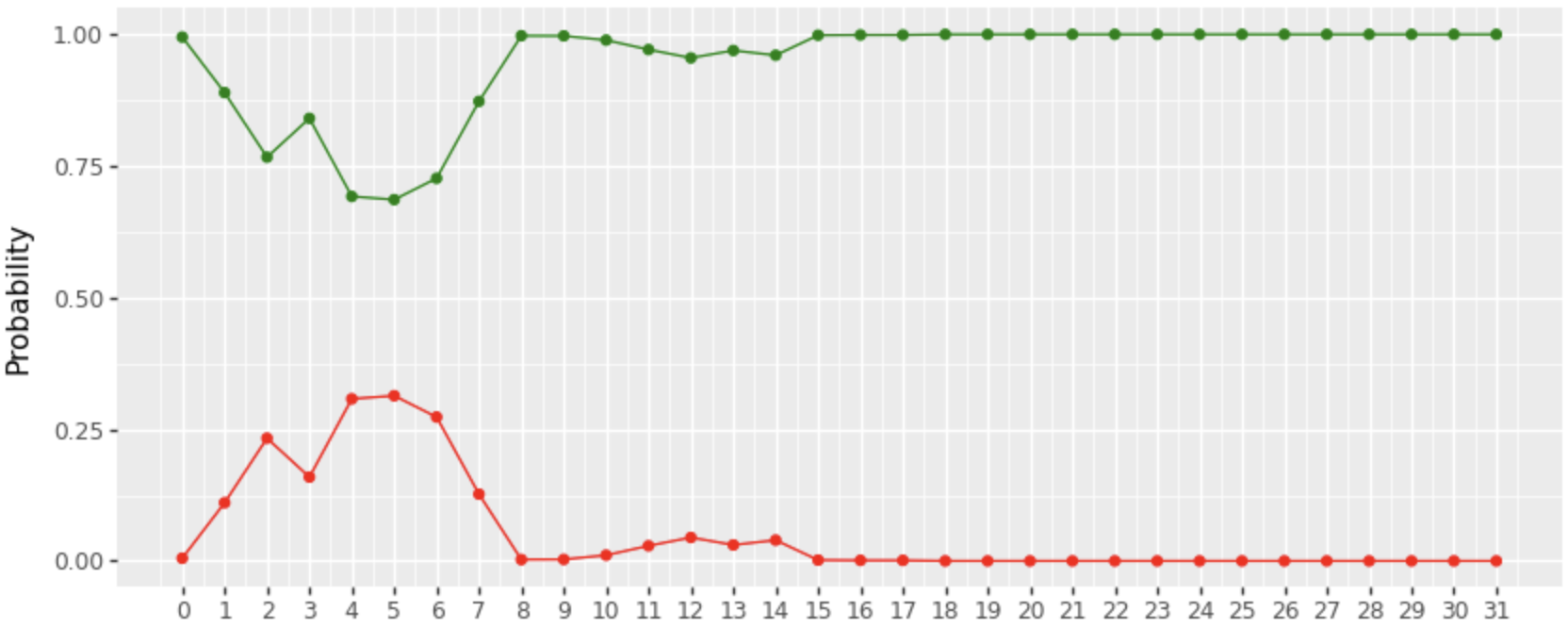}
  \caption{}
  \label{subfig:mistral_actpatching11}
\end{subfigure}\hspace{0.15cm} 
\begin{subfigure}{0.30\textwidth}
  \includegraphics[width=\linewidth]{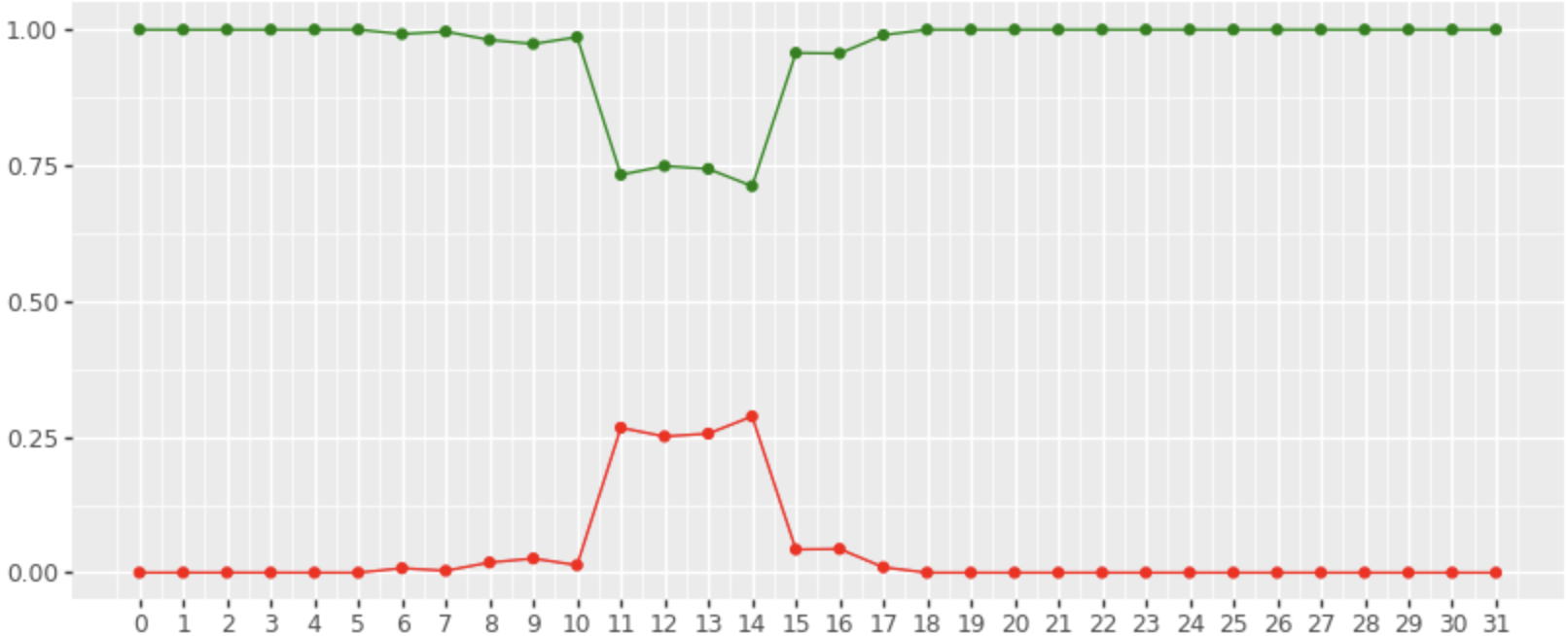}
  \caption{}
  \label{subfig:mistral_actpatching12}
\end{subfigure}\hspace{0.15cm} 
\begin{subfigure}{0.35\textwidth}
  \includegraphics[width=\linewidth]{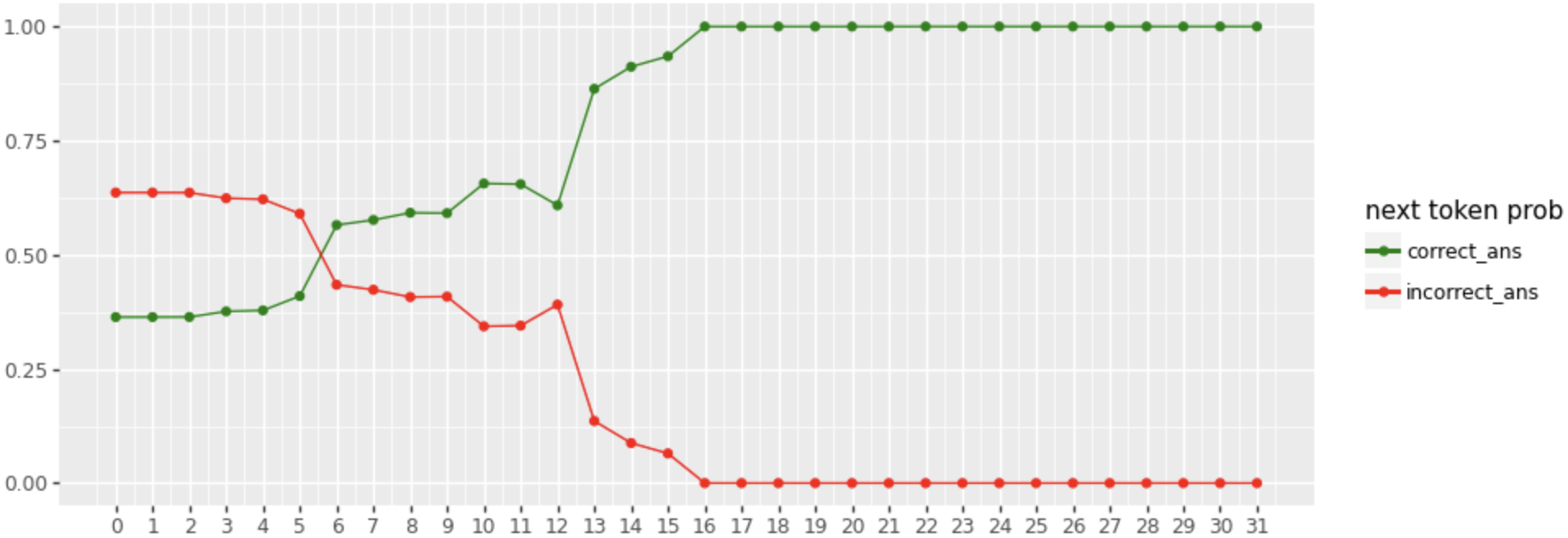}
  \caption{}
  \label{subfig:mistral_actpatching13}
\end{subfigure}

\medskip
\begin{subfigure}{0.31\textwidth}
  \includegraphics[width=\linewidth]{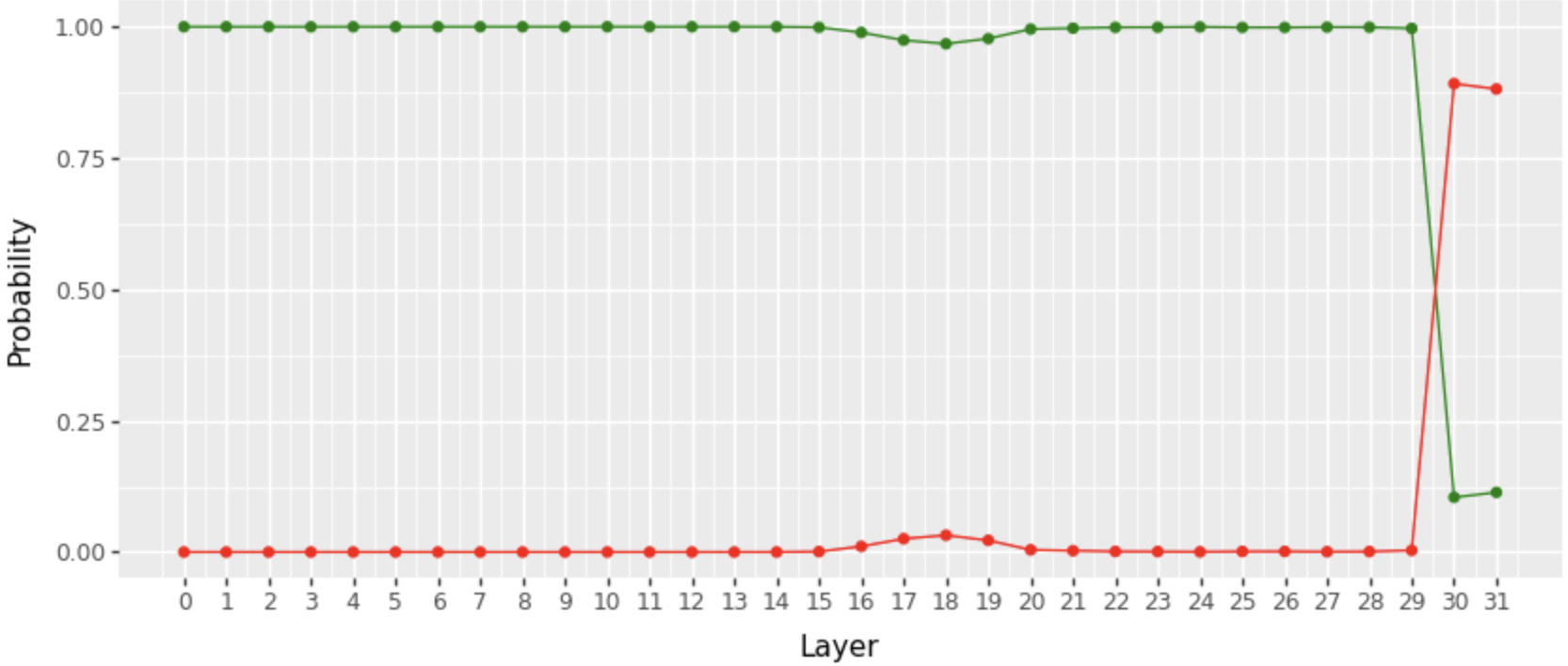}
  \caption{}
  \label{subfig:mistral_actpatching14}
\end{subfigure}\hspace{0.15cm} 
\begin{subfigure}{0.30\textwidth}
  \includegraphics[width=\linewidth]{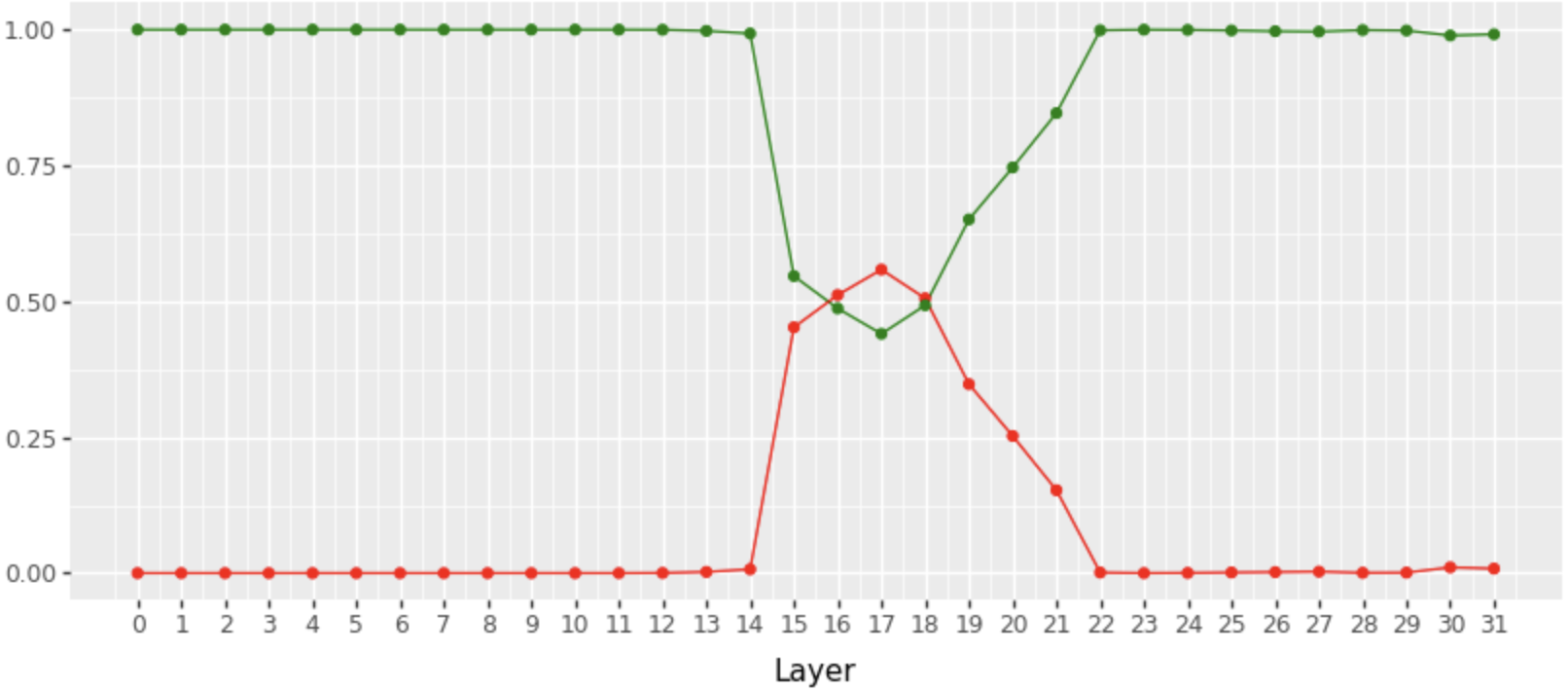}
  \caption{}
  \label{subfig:mistral_actpatching15}
\end{subfigure}\hspace{0.15cm} 
\begin{subfigure}{0.35\textwidth}
  \includegraphics[width=\linewidth]{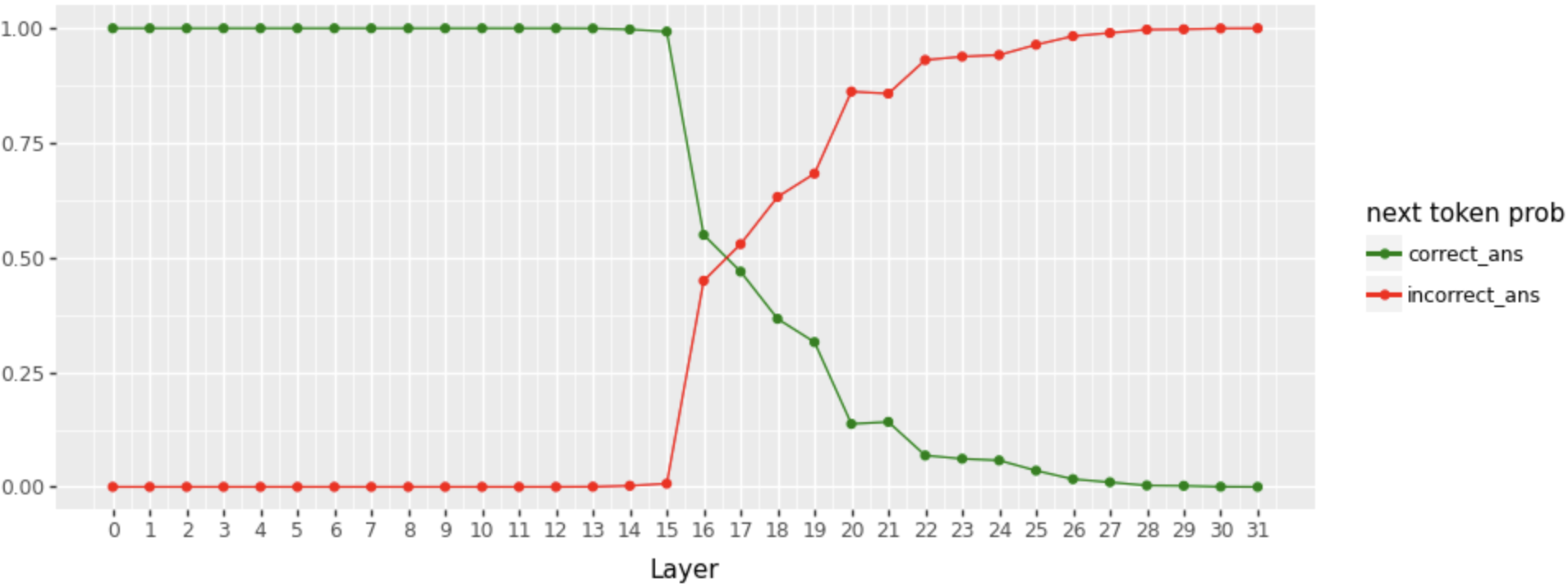}
  \caption{}
  \label{subfig:mistral_actpatching16}
\end{subfigure}
\caption{Change in probability in Mistral model from the correct answer (able or unable) to the wrong answer (unable or able) under activation patching between causation and antithesis with a positive verb. The top row shows the change in probability when patched at the token location for ``so'' and ``yet'', while the bottom row shows the same change at the last token location. (a) and (d) illustrate the effect of patching at the MLP activation. (b) and (e) show the effect of patching at the attention output, and (c) and (f) depict the effect of patching at the block output.}
\label{fig:mistral_actpatching}
\end{figure*}

\begin{figure*}[htb]
    \centering 
\begin{subfigure}{0.3\textwidth}
  \includegraphics[width=\linewidth]{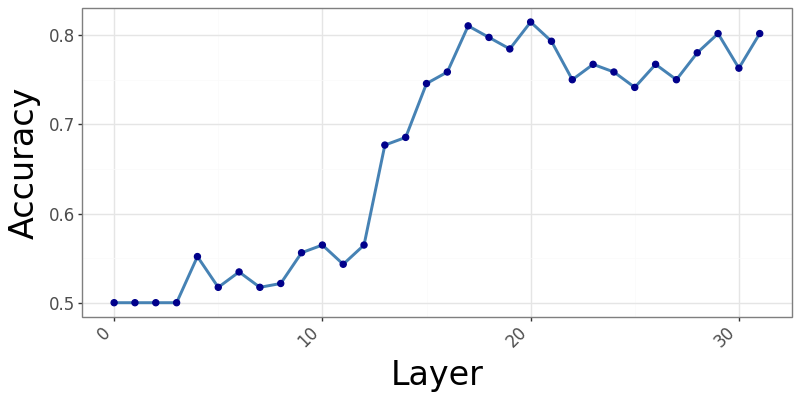}
  \caption{}
  \label{subfig:mlp0_ableunable}
\end{subfigure}\hspace{0.15cm} 
\begin{subfigure}{0.32\textwidth}
  \includegraphics[width=\linewidth]{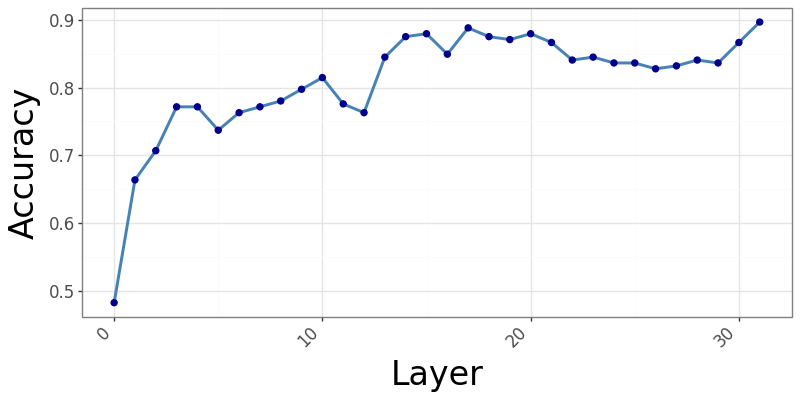}
  \caption{}
  \label{subfig:mlp0_vtype}
\end{subfigure}\hspace{0.15cm} 
\begin{subfigure}{0.32\textwidth}
  \includegraphics[width=\linewidth]{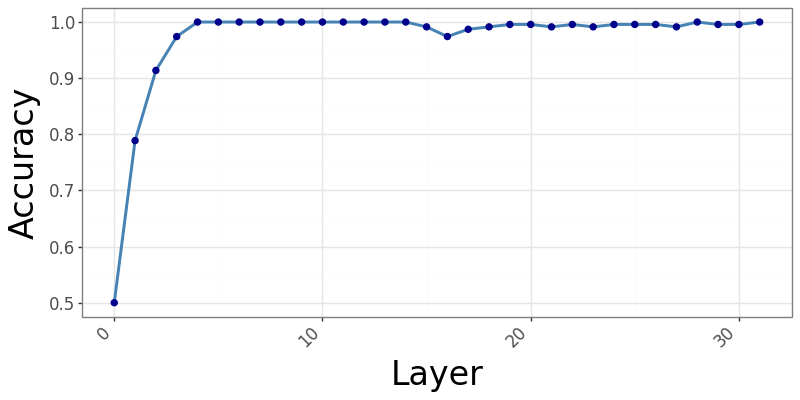}
  \caption{}
  \label{subfig:mlp0_vsense}
\end{subfigure}
\medskip

\begin{subfigure}{0.3\textwidth}
  \includegraphics[width=\linewidth]{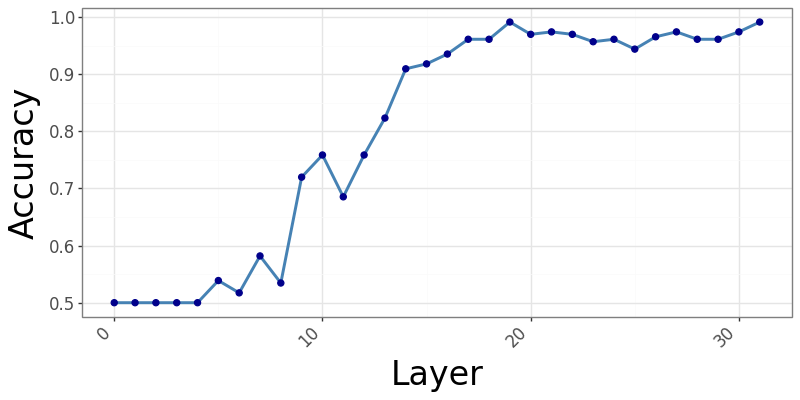}
  \caption{}
  \label{subfig:mlp_ableunable}
\end{subfigure}\hspace{0.15cm} 
\begin{subfigure}{0.32\textwidth}
  \includegraphics[width=\linewidth]{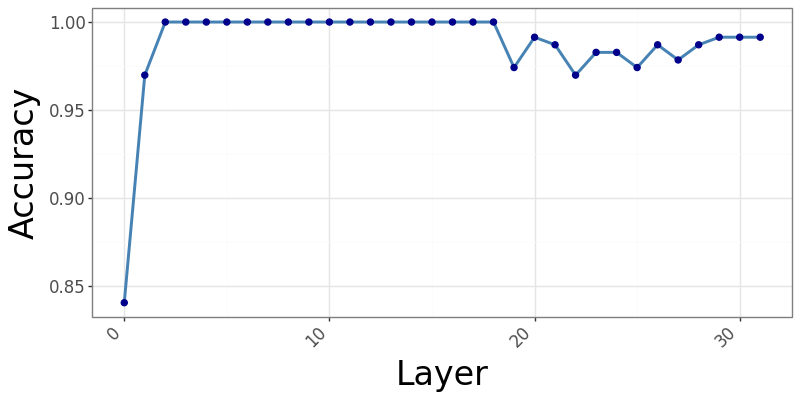}
  \caption{}
  \label{subfig:mlp_vtype}
\end{subfigure}
\begin{subfigure}{0.32\textwidth}
  \includegraphics[width=\linewidth]{images/MLP_probe_at_lasttok_overallVerbsense_All.png}
  \caption{}
  \label{subfig:mlp_vsense}
\end{subfigure}
\caption{Accuracy of logistic regression probes across MLP layers of LLaMA. Top row shows probe accuracy  at the last token position for 0-shot prompt. Bottom row shows  probe accuracy  at the last token position for 2-shots per class. (a) and (d) show classification accuracy for ``able'' vs. ``unable.'' (b) and (e) show accuracy for classifying the sentiment polarity of the satellite verb; and (c) and (f) show accuracy for predicting the sentiment of the satellite clause}
\label{fig:mlp_prob}
\end{figure*}

\begin{figure*}[htb]
    \centering 
\begin{subfigure}{0.32\textwidth}
  \includegraphics[width=\linewidth]{{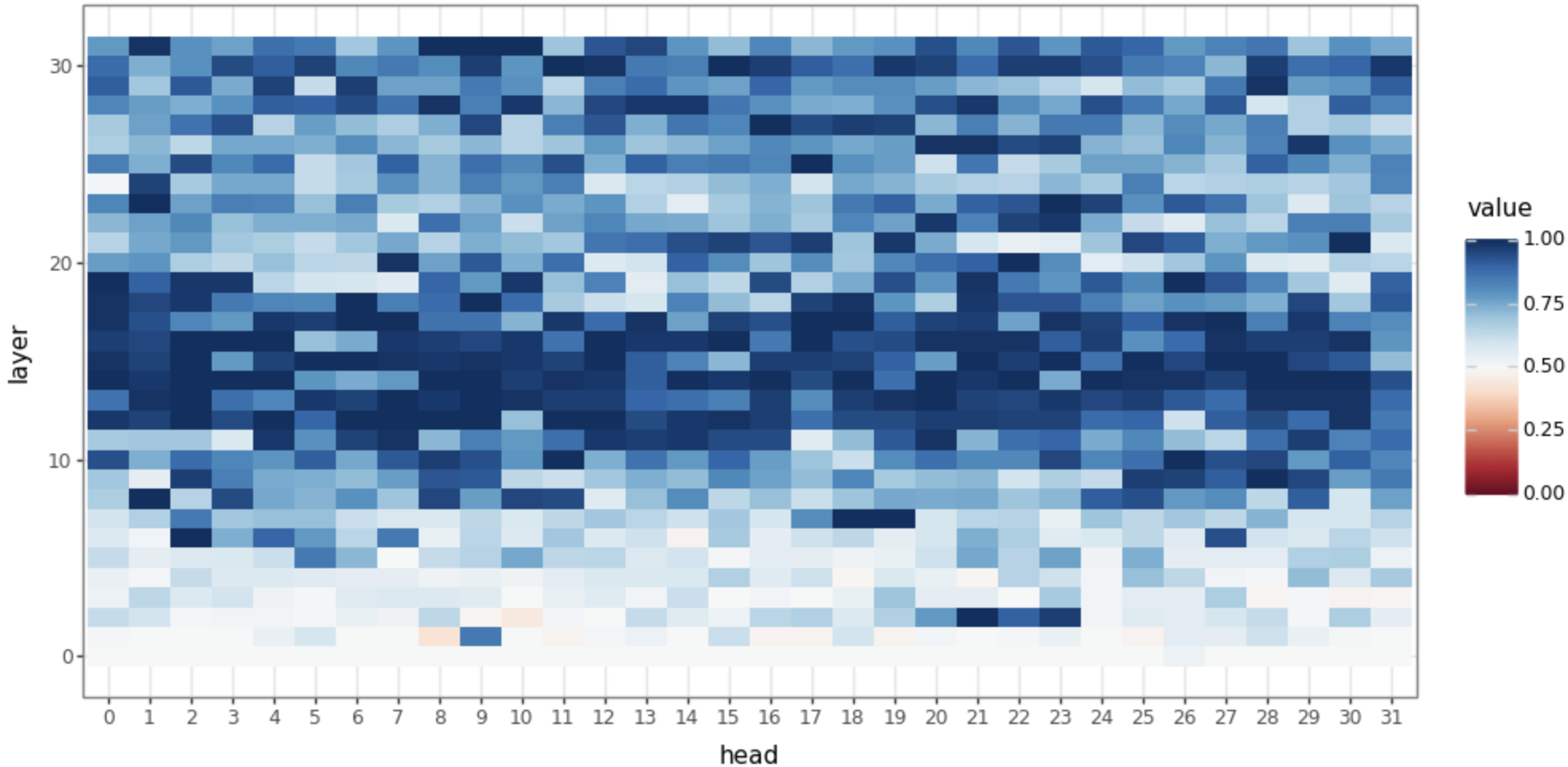}}
  \caption{}
  \label{subfig:mistral_logprob1}
\end{subfigure}\hspace{0.15cm} 
\begin{subfigure}{0.32\textwidth}
  \includegraphics[width=\linewidth]{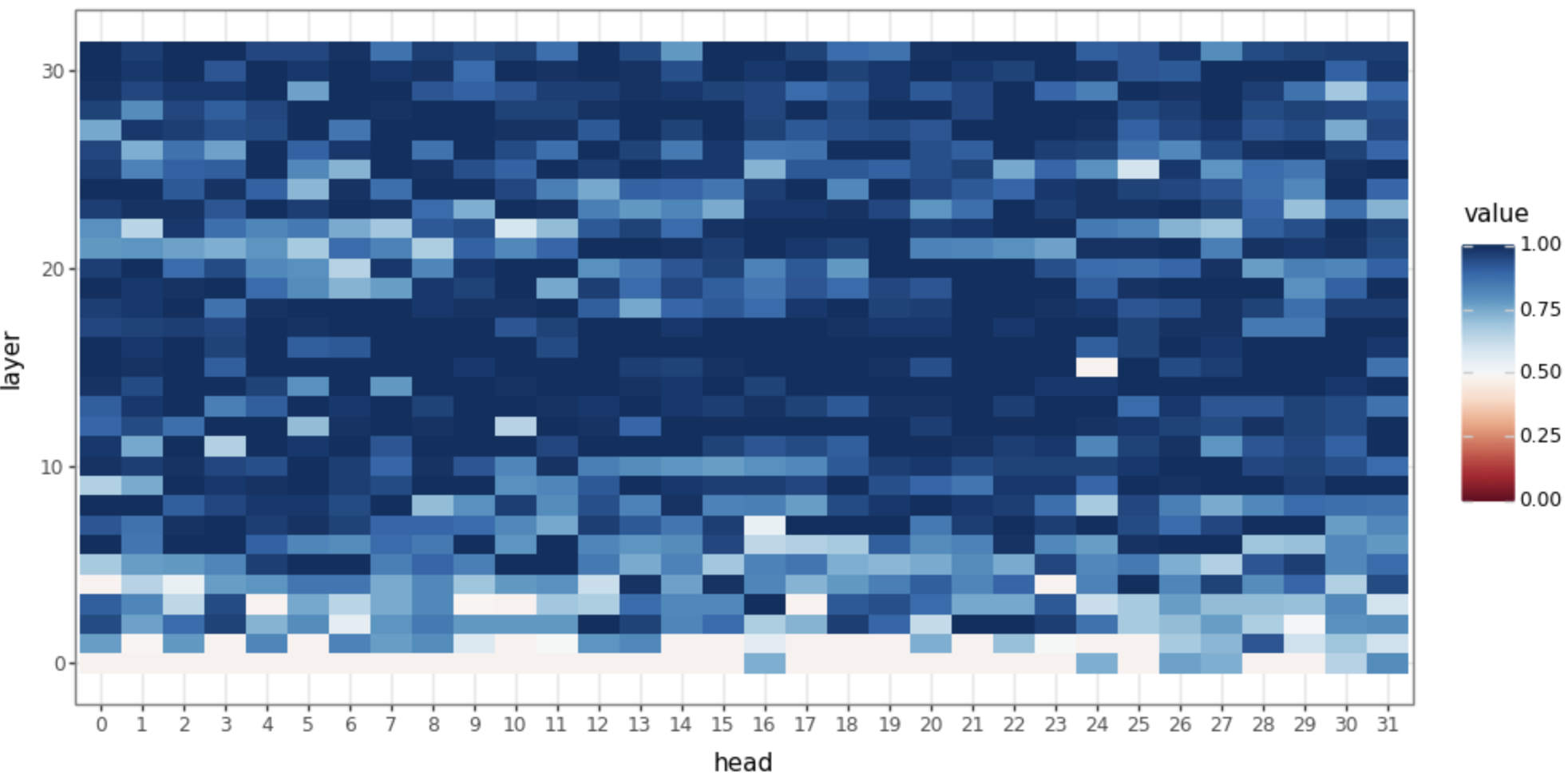}
  \caption{}
  \label{subfig:mistral_logprob2}
\end{subfigure}\hspace{0.15cm} 
\begin{subfigure}{0.32\textwidth}
  \includegraphics[width=\linewidth]{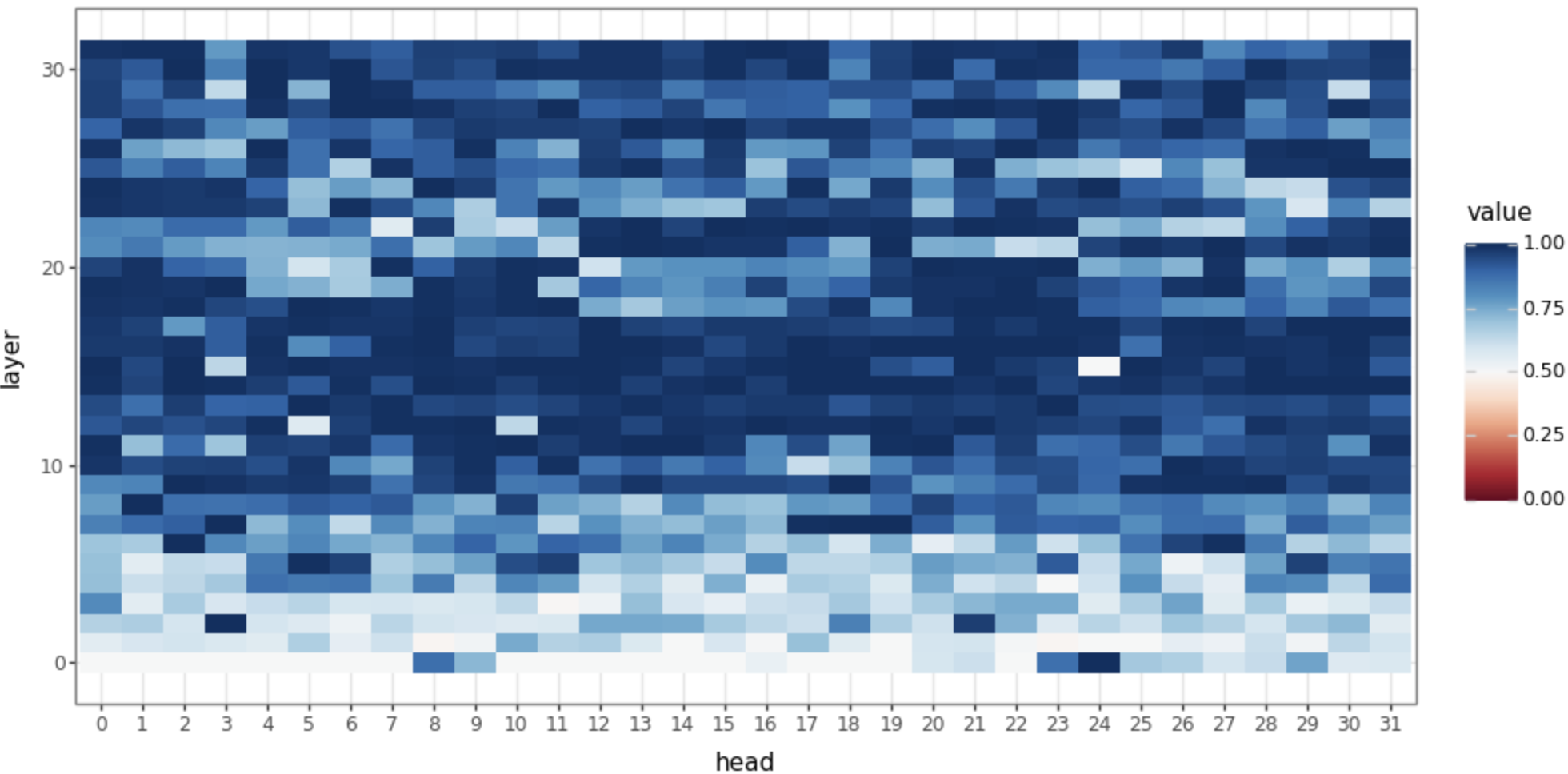}
  \caption{}
  \label{subfig:mistral_logprob3}
\end{subfigure}

\medskip

\begin{subfigure}{0.32\textwidth}
  \includegraphics[width=\linewidth]{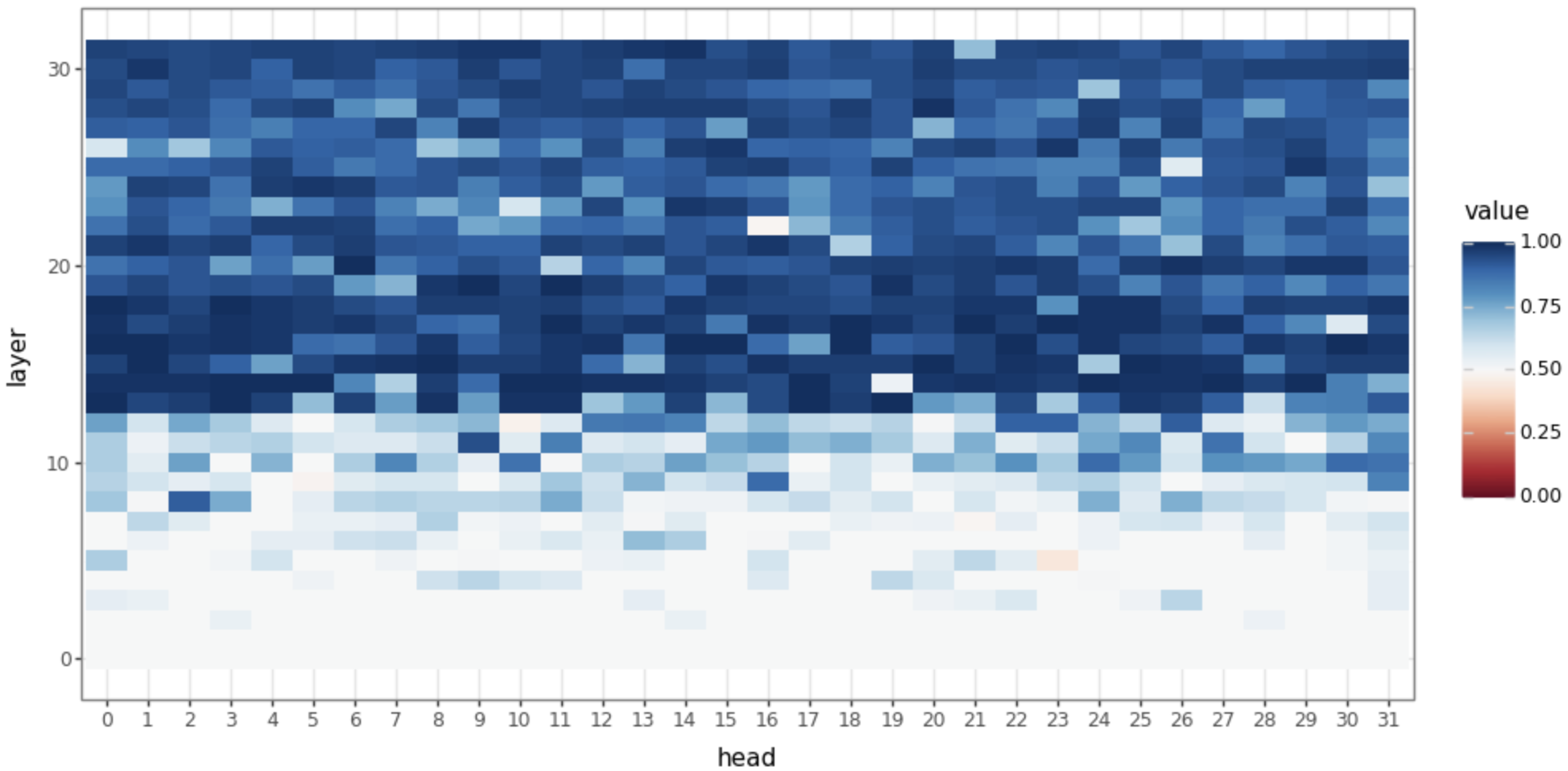}
  \caption{}
  \label{subfig:mistral_logprob4}
\end{subfigure}\hspace{0.15cm} 
\begin{subfigure}{0.32\textwidth}
  \includegraphics[width=\linewidth]{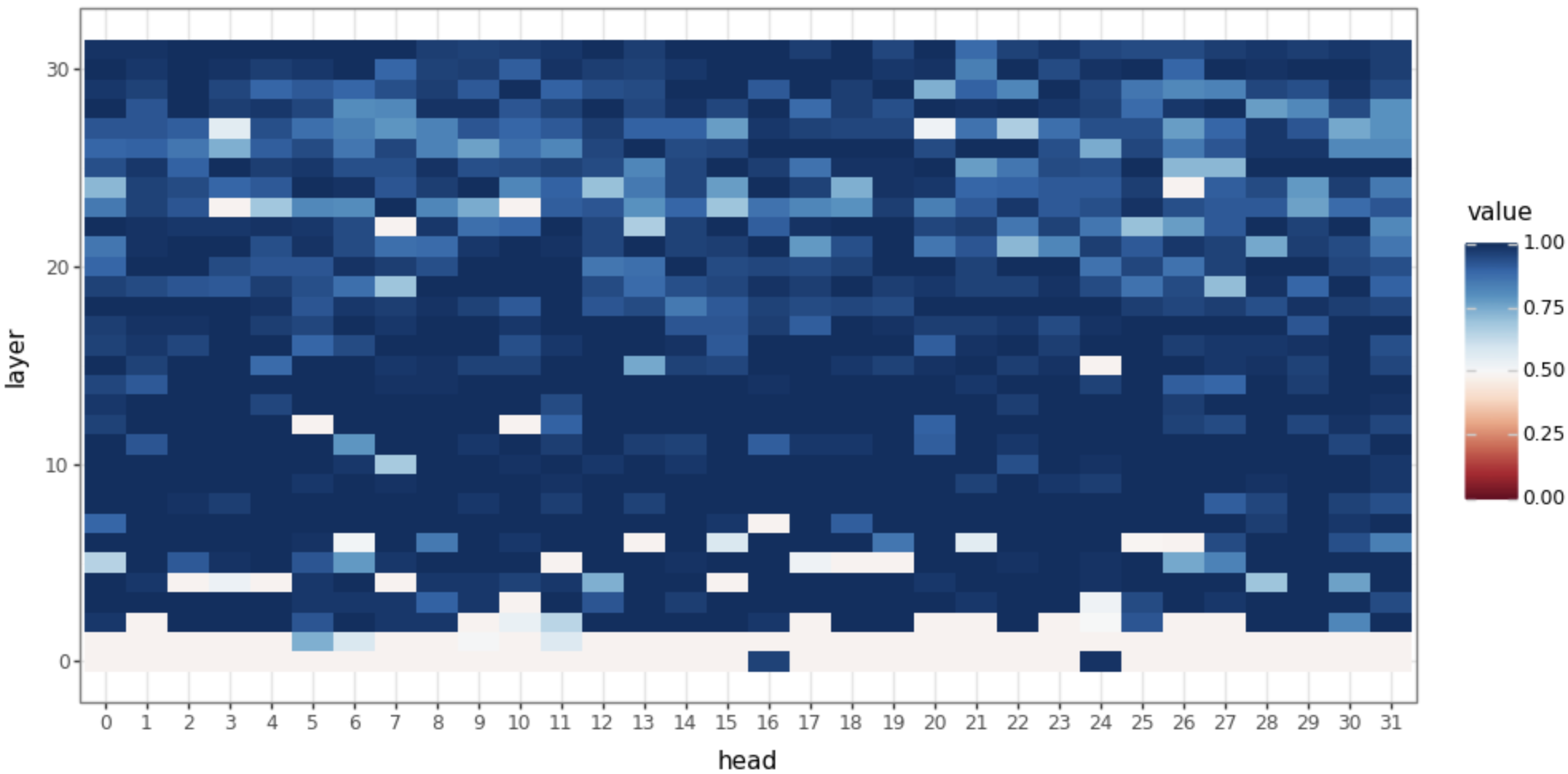}
  \caption{}
  \label{subfig:mistral_logprob5}
\end{subfigure}\hspace{0.15cm} 
\begin{subfigure}{0.32\textwidth}
  \includegraphics[width=\linewidth]{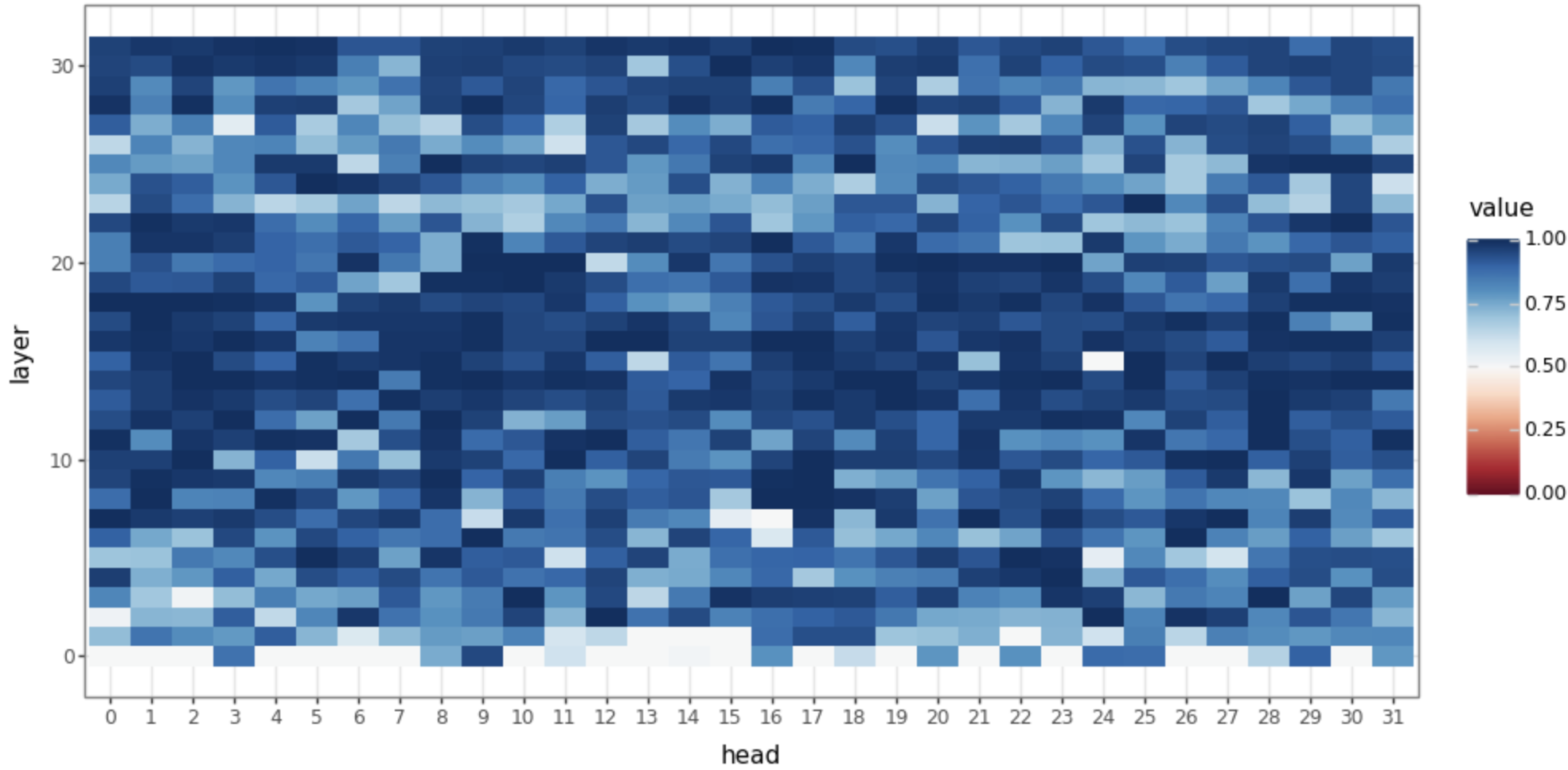}
  \caption{}
  \label{subfig:mistral_logprob6}
\end{subfigure}
\caption{Accuracy of logistic regression probes across attention heads of Mistral. The top row shows probe accuracy at the token position for ``so''\slash``yet'', while the bottom row shows accuracy at the position of the last token in the input. Subfigures (a) and (d) show classification accuracy for \emph{``able''}vs. \emph{``unable''}; (b) and (e) show accuracy for classifying the sentiment polarity of the satellite verb; and (c) and (f) show accuracy for predicting the sentiment of the satellite clause. }
\label{fig:mistral_logprob}
\end{figure*}

\end{document}